\documentclass[10pt,twocolumn,letterpaper]{article}

\usepackage{cvpr}

\makeatletter
\@namedef{ver@everyshi.sty}{}
\makeatother

\usepackage{graphicx}
\usepackage{amsmath}
\usepackage{amssymb}
\usepackage{booktabs}
\usepackage{comment}
\usepackage{color}
\usepackage{tabularx}
\usepackage{colortbl}
\usepackage{tabu}
\usepackage{multirow}
\usepackage{setspace}
\usepackage[font=small,labelsep=period]{caption}

\usepackage[moderate]{savetrees}
\usepackage[usenames,dvipsnames]{xcolor}
\usepackage{tikz}
\usepackage{pgfplots}
\usepackage{pgfkeys}
\usepackage{xcolor}

\usepackage[accsupp]{axessibility} %

\makeatletter
\renewcommand{\fnum@figure}{Figure \thefigure}
\makeatother

\newcolumntype{Y}{>{\centering\arraybackslash}X}
\newcolumntype{R}{>{\raggedleft\arraybackslash}X}
\newcolumntype{L}{>{\raggedright\arraybackslash}X}

\newcommand{\mytilde}{\raise.17ex\hbox{$\scriptstyle\sim$}}

\definecolor{bblue}{rgb}{0.0,0.2,0.8}
\definecolor{ccol}{rgb}{0.91,0.91,0.91}

\definecolor{avgcol}{rgb}{1.0,0.91,0.722}
\definecolor{arcol}{rgb}{0.765,0.878,0.812}
\definecolor{timecol}{rgb}{0.941,0.749,0.737}

\usepackage{xspace}

\newcommand\customparagraph[1]{\vspace{0.7em}\noindent\textbf{#1}}

\def\addlegendimage{\csname pgfplots@addlegendimage\endcsname}

\usepackage[pagebackref=true,breaklinks=true,colorlinks=true,urlcolor=bblue,bookmarks=false,citecolor=bblue]{hyperref}

\usepackage[capitalize]{cleveref}
\crefname{section}{Sec.}{Secs.}
\Crefname{section}{Section}{Sections}
\Crefname{table}{Table}{Tables}
\crefname{table}{Tab.}{Tabs.}

\def\cvprPaperID{11}
\def\confName{CVPR}
\def\confYear{2024}

\begin{document}

\title{\vspace{-1ex}BOP Challenge 2023 on Detection, Segmentation and\\Pose Estimation of Seen and Unseen Rigid Objects \vspace{-1ex}}

\newcommand{\namesep}{\hspace{0.8em}}
\author{
 Tomas~Hodan$^{1}$\namesep
 Martin Sundermeyer$^{2}$\namesep
 Yann Labb{\'e}$^{1}$\namesep
 Van Nguyen Nguyen$^{3}$\namesep
 Gu Wang$^{4}$\\
 Eric Brachmann$^{5}$\namesep
 Bertram Drost$^{6}$\namesep
 Vincent Lepetit$^{3}$\namesep
 Carsten Rother$^{7}$\namesep
 Jiri~Matas$^{8}$\vspace{0.7em} \\
 {\normalsize
     {$^{1}$Meta}\namesep
     {$^{2}$Google}\namesep
     {$^{3}$ENPC} \namesep
     {$^{4}$Tsinghua University} \namesep
     {$^{5}$Niantic}\namesep
     {$^{6}$MVTec}\namesep
     {$^{7}$Heidelberg University}\namesep
     {$^{8}$CTU in Prague}
 }
}

\maketitle
\begin{abstract}
 \vspace{-1ex}
We present the evaluation methodology, datasets and results of the BOP Challenge 2023, the fifth in a series of public competitions organized to capture the state of the art in model-based 6D object pose estimation from an RGB/RGB-D image and related tasks. Besides the three tasks from 2022 (2D detection, 2D segmentation, and 6D localization of objects seen during training), the 2023 challenge introduced new variants of these tasks focused on objects unseen during training. In the new tasks, methods were required to learn new objects during a short onboarding stage (max 5 minutes, 1 GPU) from provided 3D object models. The best 2023 method for 6D localization of unseen objects (GenFlow) notably reached the accuracy of the best 2020 method for seen objects (CosyPose), although being noticeably slower. The best 2023 method for seen objects (GPose) achieved a moderate accuracy improvement but a significant 43\% run time improvement compared to the best 2022 counterpart (GDRNPP). Since 2017, the accuracy of 6D localization of seen objects has improved by more than 50\% (from 56.9 to 85.6 AR$_C$). The online evaluation system stays open and is available at:~\texttt{\href{http://bop.felk.cvut.cz/}{bop.felk.cvut.cz}}.

\end{abstract}

\vspace{-2.0ex}
\section{Introduction}

The BOP Challenge 2023 was the fifth in a series of public challenges that are part of the BOP\footnote{BOP stands for Benchmark for 6D Object Pose Estimation~\cite{hodan2018bop}.} project, which aims to continuously record and report the state of the art in estimating the 6D object pose (3D translation and 3D rotation) and related tasks such as 2D object detection and segmentation.
Results of the previous editions of the challenge from 2017, 2019, 2020, and 2022 were published in~\cite{hodan2018bop, hodan2019bop, hodan2020bop, sundermeyer2022bop}.

Participants of the 2023 challenge were competing on six tasks. Besides the three tasks from 2022 (model-based 2D object detection, 2D object segmentation and 6D object localization of objects seen during training), the 2023 challenge introduced new variants of these tasks focused on \emph{objects unseen during training}. In the new tasks, methods were required to adapt to novel 3D object models during a short object onboarding stage (max 5 min per object, 1 GPU), and then recognize the objects in images from diverse environments. Such methods are of high practical relevance as they do not require expensive data generation and training for every new object, which is typically required by most existing methods for seen objects and severely limits their scalability. The introduction of the new tasks was encouraged by the recent breakthroughs in foundation models and their impressive few-shot learning capabilities.
\begin{figure}
\centering
\begin{tikzpicture}
    \tikzstyle{every node}=[font=\small]
    \begin{axis}[
        width=9.cm,
        height=5.5cm,
        font=\footnotesize,
        xlabel=Run time per image (s),
        xlabel style={at={(axis description cs:0.5,-.08)},anchor=south},
        ylabel=Accuracy ($\text{AR}_C$),
        ylabel style={at={(axis description cs:0.12,.5)},anchor=south}, %
        ymin=50,
        ymax=95,
        ytick={55, 65, 75, 85, 95},
        xmin=0,
        xmax=55,
        xtick={0, 10, 20, 30, 40, 50},
        label style={font=\small},
        tick label style={font=\small},
        legend pos=south east,
        grid=major,
        legend style={nodes={scale=1, transform shape}},
        legend cell align={left}
    ]

    \addlegendimage{only marks, mark=x, color=blue, mark size=4pt, line width=1.5pt} %
    \addlegendentry{Seen objects}
    
    \addlegendimage{only marks, mark=*, color=Orange, mark size=3pt}
    \addlegendentry{Unseen objects}
    
    \draw plot[mark=x, color=blue, mark size=2pt,mark options={color=blue,mark size=4pt,line width=1.5pt}] coordinates {(axis cs: 2.67,85.6)};
    \node[anchor=west] at (axis cs:2.77,87.6) {\small GPose (2023)};

    \draw plot[mark=x, color=Orange, mark size=2pt,mark options={color=blue,line width=1.5pt, mark size=4pt}] coordinates {(axis cs: 6.263,83.7)};
    \node[anchor=west] at (axis cs:6.563,81.7) {GDRNPP (2022)};

    \draw plot[mark=x, color=blue, mark size=2pt,mark options={color=blue,line width=1.5pt,mark size=4pt}] coordinates {(axis cs: 13.74,69.8)};
    \node[anchor=west] at (axis cs:13.94,71.8) {CosyPose (2020)};

    \draw plot[mark=x, color=blue, mark size=2pt,mark options={color=blue,mark size=4pt, line width=1.5pt}] coordinates {(axis cs: 3.220,56.9)};
    \node[anchor=west] at (axis cs:3.420,58.9) {Vidal \etal\ (2017)};

    \draw plot[mark=*, color=Orange, mark size=2pt,mark options={color=Orange,mark size=3pt}] coordinates {(axis cs: 34.57,67.4)};
    \node[anchor=west] at (axis cs:34.77,69.4) {GenFlow (2023)};

    \end{axis}
\end{tikzpicture}
\vspace{-0.35cm}
 \caption{\textbf{Progress in model-based 6D object localization (2017--2023).} Shown is the accuracy and run time of the top performing RGB-D methods on the seven core BOP datasets.
 The dominance of methods based on point-pair features~\cite{drost2010model}, represented by Vidal~\etal~\cite{vidal2018method} in 2017, was ended by the learning-based CosyPose~\cite{labbe2020cosypose} in 2020 for the price of a significantly higher run time. In 2022, GDRNPP~\cite{Wang_2021_GDRN,liu2022gdrnpp_bop} dramatically improved both accuracy and run time. Finally, in 2023, GPose~\cite{gpose2023} brought the run time back to the 2017 level while further improving the accuracy. The field has come a long way since 2017 -- the accuracy has improved by more than 50\% (from 56.9 to 85.6 AR$_C$).
 GenFlow~\cite{genflow}, the best method for the newly introduced task of 6D localization of \emph{unseen objects} (objects not seen during training), reaches the accuracy of CosyPose, the best 2020 method for \emph{seen objects}, while its run time awaits improvements.
 \vspace{1.0ex}
 }
\label{fig:teaser}
\end{figure}
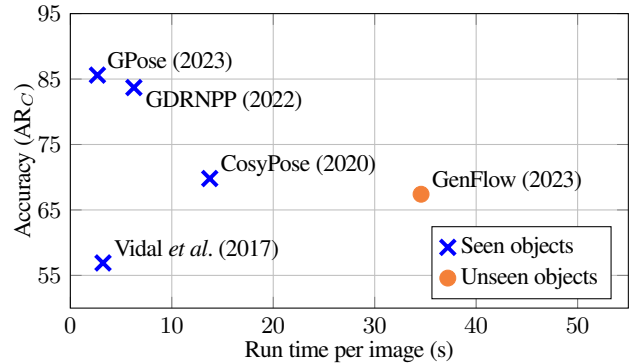

The challenge primarily focuses on the practical scenario where no real images are available at training/onboarding time, only the 3D object models and images synthesized using the models. While capturing real images of objects under various conditions and annotating the images with 6D object poses requires a significant human effort~\cite{hodan2017tless}, the 3D models are either available before the physical objects, which is often the case for manufactured objects, or can be reconstructed at an admissible cost.
Approaches for reconstructing 3D models of opaque, matte and moderately specular objects are established~\cite{newcombe2011kinectfusion,reizenstein2021common} and promising approaches for transparent and highly specular objects are emerging~\cite{wu2018full,Munkberg_2022_CVPR,hasselgren2022shape,verbin2022ref}.

In the 2019 challenge, methods using the depth image channel were mostly based on point pair features (PPF's)~\cite{drost2010model} and clearly outperformed methods relying only on the RGB channels, all of which were based on deep neural networks (DNN's). DNN-based methods need large amounts of annotated training images, which had been typically obtained by OpenGL rendering of the 3D object models on random backgrounds~\cite{kehl2017ssd,hinterstoisser2017pre}. However, as suggested in~\cite{hodan2019photorealistic}, the evident domain gap between these ``render\;\&\;paste'' training images and real test images
limits the potential of the DNN-based methods. To reduce the gap between the synthetic and real domains and thus to bring fresh air to the DNN world, we joined the development of BlenderProc\footnote{\href{https://github.com/DLR-RM/BlenderProc/blob/main/README_BlenderProc4BOP.md}{\texttt{github.com/DLR-RM/BlenderProc}}}~\cite{denninger2019blenderproc,denninger2020blenderproc}, an open-source, physically-based renderer (PBR).
For the 2020 challenge, we then provided participants with 350K PBR training images (see~\cite{hodan2020bop} for examples), which helped the DNN-based methods to achieve noticeably higher accuracy and to finally catch up with the PPF-based methods. In the 2022 challenge, DNN-based methods for 6D object localization already clearly outperformed PPF-based methods in both accuracy and speed, with the performance gains coming mostly from advances in network architectures and training schemes.

Remarkably, RGB methods from 2022 surpassed RGB-D methods from 2020, the performance gap between methods trained only on PBR images and methods trained also on real images noticeably shrank, and some methods started training on the depth image channel in addition to the RGB channels. 
In 2022, we started evaluating also the tasks of 2D object detection and 2D object segmentation, to address the design of the majority of recent object pose estimation methods, which start by detecting/segmenting objects and then estimate their poses from the predicted image regions. Evaluating the detection/segmentation and pose estimation stages separately enabled a better understanding of the progress in object pose estimation.

In 2023, we introduced three more practical tasks focused on unseen objects, \ie the target objects are not seen during training and need to be onboarded with limited resources (max 5 minutes on 1 GPU). While similar tasks have been considered in the literature~\cite{megapose, nguyen2022template, shugurov2022osop}, direct comparison of methods has been difficult due to variations in the detection stage and the used training data. To address this situation, we proposed a unified evaluation framework utilizing an open-source detection method and a large-scale training dataset. Specifically, CNOS~\cite{nguyen2023cnos}, a model-based method for detecting/segmenting unseen objects that outperforms Mask-RCNN~\cite{he2017mask}, was employed as the default method for 2D detection and segmentation. As the training dataset, we used synthetic training data from MegaPose~\cite{megapose}. Methods were not required but encouraged (via dedicated awards) to use these unified solutions.

The best 2023 method for 6D localization of unseen objects (GenFlow~\cite{genflow}) reached the accuracy of the best 2020 method for seen objects (CosyPose~\cite{labbe2020cosypose}). Despite being noticeably slower, this is an impressive result considering that the target objects are onboarded in a short time, which is several orders of magnitude shorter than a typical training process of methods trained for specific objects. The best 2023 method for seen objects (GPose~\cite{gpose2023}) achieves a moderate accuracy improvement and a significant 42.6\% run time improvement compared to the best 2022 counterpart (GDRNPP~\cite{Wang_2021_GDRN,liu2022gdrnpp_bop}).

Sec.~\ref{sec:methodology} of this report defines the evaluation methodology, Sec.~\ref{sec:datasets} introduces datasets, Sec.~\ref{sec:evaluation} describes the experimental setup and analyzes the results, Sec.~\ref{sec:awards} presents the awards of the BOP Challenge 2023, and Sec.~\ref{sec:conclusion} concludes the report.

\section{Challenge tasks} \label{sec:methodology}

Methods are evaluated on the task of model-based 6D localization on seen objects (as in 2019, 2020 and 2022~\cite{sundermeyer2022bop}), on the tasks of model-based 2D detection and 2D segmentation of seen objects (as in 2022~\cite{sundermeyer2022bop}), and on variants of these tasks focused on objects unseen during training, which were introduced in 2023.
All six tasks are defined below, together with accuracy scores that are used to compare methods. Participants could submit their results to any of the six tasks. Note that although all BOP datasets currently include RGB-D images (Sec.~\ref{sec:datasets}), a method may have used any of the image channels.

\subsection{Task 1: 6D localization of seen objects}
\label{sec:task1}

The definition of this task is the same since 2019, which enables direct comparison across the years\footnote{See Sec.~A.1 in~\cite{hodan2020bop} for a discussion on why the methods are evaluated on 6D object localization instead of 6D object detection, where no prior information about the visible object instances is provided~\cite{hodan2016evaluation}.}.

\customparagraph{Training input:}
At training time, a method is provided a set of RGB-D training images showing objects annotated with ground-truth 6D poses, and 3D mesh models of the objects (typically with a color texture).
A 6D pose is defined by a matrix $\textbf{P} = [\mathbf{R} \, | \, \mathbf{t}]$, where $\mathbf{R}$ is a 3D rotation matrix, and $\mathbf{t}$ is a 3D translation vector. The matrix $\textbf{P}$ defines a rigid transformation from the 3D space of the object model to the 3D space of the camera.

\customparagraph{Test input:}
At test time, the method is given an RGB-D image unseen during training and a list $L = [(o_1, n_1),$ $\dots,$ $(o_m, n_m)]$, where $n_i$ is the number of instances of object $o_i$ visible in the image.
In 2023, methods could use provided default detections (results of GDRNPPDet\_PBRReal, the best 2D detection method from 2022 for Task 2).

\customparagraph{Test output:} The method produces a list $E=[E_1,$$\dots,$$E_m]$, where $E_i$ is a list of $n_i$ pose estimates with confidences for instances of object $o_i$.

\customparagraph{Evaluation methodology:}
The error of an estimated pose \wrt the ground-truth pose is calculated by three pose-error functions (see Sec.~2.2 of~\cite{hodan2020bop} for details): (1) VSD (Visible Surface Discrepancy) which treats indistinguishable poses as equivalent by considering only the visible object part, (2) MSSD (Maximum Symmetry-Aware Surface Distance) which considers a set of pre-identified global object symmetries and measures the surface deviation in 3D, (3) MSPD (Maximum Symmetry-Aware Projection Distance) which considers the object symmetries and measures the perceivable deviation.

An estimated pose is considered correct \wrt a pose-error function~$e$, if $e < \theta_e$, where $e \in \{\text{VSD}, \text{MSSD}, \text{MSPD}\}$ and $\theta_e$ is the threshold of correctness. The fraction of annotated object instances for which a correct pose is estimated is referred to as Recall. The Average Recall \wrt a function~$e$, denoted as $\text{AR}_e$, is defined as the average of the Recall rates calculated for multiple settings of the threshold $\theta_e$ and also for multiple settings of a misalignment tolerance $\tau$ in the case of $\text{VSD}$. The accuracy of a method on a dataset $D$ is measured by: $\text{AR}_D = (\text{AR}_\text{VSD} + \text{AR}_{\text{MSSD}} + \text{AR}_{\text{MSPD}}) \, / \, 3$, which is calculated over estimated poses of all objects from $D$. The overall accuracy on the core datasets is measured by $\text{AR}_C$ defined as the average of the per-dataset $\text{AR}_D$ scores (see Sec.~2.4 of~\cite{hodan2020bop} for details)\footnote{When calculating AR$_C$, scores are not averaged over objects before averaging over datasets, which is done when calculating $\text{AP}_C$ (Sec.~\ref{sec:task2}) to comply with the original COCO evaluation methodology~\cite{lin2014microsoft}.}.

\subsection{Task 2: 2D detection of seen objects}
\label{sec:task2}

\noindent\textbf{Training input:}
At training time, a method is provided a set of RGB-D training images showing objects annotated with ground-truth 2D bounding boxes. The boxes are \emph{amodal}, \ie, covering the whole object silhouette, including the occluded parts. The method can use the 3D mesh models that are available for the objects (\eg, to synthesize extra training images).

\customparagraph{Test input:}
At test time, the method is given an RGB-D image unseen during training that shows an arbitrary number of instances of an arbitrary number of objects, with all objects being from one specified dataset (\eg YCB-V \cite{xiang2017posecnn}). No prior information about the visible object instances is provided.

\customparagraph{Test output:}
The method produces a list of object detections with confidences, with each detection defined by an \emph{amodal} 2D bounding box.

\customparagraph{Evaluation methodology:}
Following the evaluation methodology from the COCO 2020 Object Detection Challenge~\cite{lin2014microsoft}, the detection accuracy is measured by the Average Precision (AP). Specifically, a per-object $\text{AP}_O$ score is calculated by averaging the precision at multiple Intersection over Union (IoU) thresholds: $[0.5, 0.55, \dots , 0.95]$. The accuracy of a method on a dataset $D$ is measured by $\text{AP}_D$ calculated by averaging per-object $\text{AP}_O$ scores, and the overall accuracy on the core datasets (Sec.~\ref{sec:datasets}) is measured by $\text{AP}_C$ defined as the average of the per-dataset $\text{AP}_D$ scores.
Analogous to the 6D localization task, only annotated object instances for which at least $10\%$ of the projected surface area is visible need to be detected. Correct predictions for instances that are visible from less than $10\%$ are filtered out and not counted as false positives. 
Up to $100$ predictions per image with the highest confidences are considered.

\subsection{Task 3: 2D segmentation of seen objects}
\label{sec:task3}

\noindent\textbf{Training input:}
At training time, a method is provided a set of RGB-D training images showing objects that are annotated with ground-truth 2D binary masks. The masks are \emph{modal}, \ie, covering only the visible object parts. The method can also use 3D mesh models that are available for the objects.

\customparagraph{Test input:}
At test time, the method is given an RGB-D image unseen during training that shows an arbitrary number of instances of an arbitrary number of objects, with all objects being from one specified dataset (\eg YCB-V). No prior information about the visible object instances is provided.

\customparagraph{Test output:}
The method produces a list of object segmentations with confidences, with each segmentation defined by a \emph{modal} 2D binary mask.

\customparagraph{Evaluation methodology:}
As in Task 2, with the only difference being that IoU is calculated on masks instead of bounding boxes.

\subsection{Task 4: 6D localization of unseen objects}
\label{sec:task4}

\noindent\textbf{Training input:}
At training time, a method is provided a set of RGB-D training images showing training objects annotated with ground-truth 6D poses, and 3D mesh models of the objects (typically with a color texture). The 6D object pose is defined as in Task 1. The method can use 3D mesh models that are available for the training objects.

\customparagraph{Object-onboarding input:}
The method is provided 3D mesh models of test objects that were not seen during training. To onboard each object (\eg to render images/templates or fine-tune a neural network), the method can spend up to 5 minutes of the wall-clock time on a computer with a single GPU. The time is measured from the point right after the raw data (\eg 3D mesh models) is loaded to the point when the object is onboarded. The method can render images of the 3D object models but cannot use any real images of the objects for onboarding. The object representation (which may be given by a set of templates, a machine-learning model, \etc) needs to be fixed after onboarding (it cannot be updated on test images).

\customparagraph{Test input:}
At test time, the method is given an RGB-D image unseen during training and a list $L = [(o_1, n_1),$ $\dots,$ $(o_m, n_m)]$, where $n_i$ is the number of instances of object $o_i$ visible in the image. In 2023, the method can use provided default detections/segmentations produced by CNOS~\cite{nguyen2023cnos}.

\customparagraph{Test output:}
As in Task 1.

\customparagraph{Evaluation methodology:}
As in Task 1.

\subsection{Task 5: 2D detection of unseen objects}
\label{sec:task5}

\noindent\textbf{Training input:}
At training time, a method is provided a set of RGB-D training images showing training objects that are annotated with ground-truth 2D bounding boxes. The boxes are \emph{amodal}, \ie, covering the whole object silhouette including the occluded parts. The method can also use 3D mesh models that are available for the training objects.

\customparagraph{Object-onboarding input:}
As in Task 4.

\customparagraph{Test input:}
At test time, the method is given an RGB-D image unseen during training that shows an arbitrary number of instances of an arbitrary number of test objects, with all objects being from one specified dataset (\eg YCB-V). No prior information about the visible object instances is provided.

\customparagraph{Test output:}
As in Task 2.

\customparagraph{Evaluation methodology:}
As in Task 2.

\subsection{Task 6: 2D segmentation of unseen objects}
\label{sec:task6}

\noindent\textbf{Training input:}
At training time, a method is provided a set of RGB-D training images showing training objects that are annotated with ground-truth 2D binary masks. The masks are \emph{modal}, \ie, covering only the visible object parts. The method can also use 3D mesh models that are available for the training objects.

\customparagraph{Object-onboarding input:}
As in Task 4.

\customparagraph{Test input:}
As in Task 5.

\customparagraph{Test output:}
As in Task 3.

\customparagraph{Evaluation methodology:}
As in Task 3.

\begin{figure}[t!]
\begin{center}

\begingroup
\scriptsize

\renewcommand{\arraystretch}{0.9}

\begin{tabular}{ @{}c@{ } @{}c@{ } @{}c@{ } @{}c@{ } }
LM~\cite{hinterstoisser2012accv} & LM-O*~\cite{brachmann2014learning} & T-LESS*~\cite{hodan2017tless} & ITODD*~\cite{drost2017introducing} \vspace{0.5ex} \\
\includegraphics[width=0.243\columnwidth]{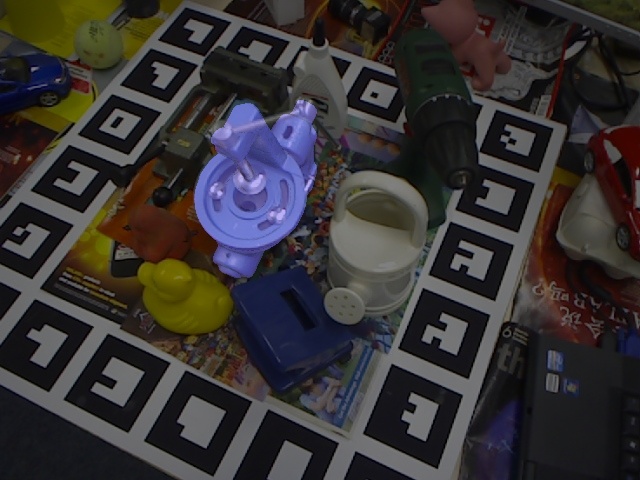} &
\includegraphics[width=0.243\columnwidth]{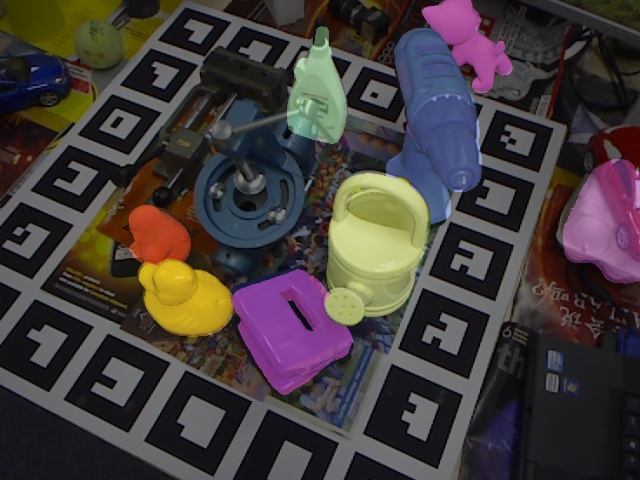} &
\includegraphics[width=0.243\columnwidth]{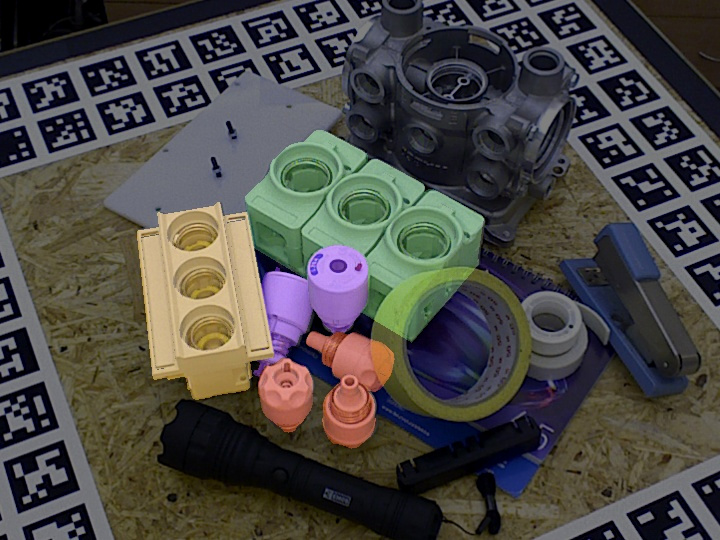} &
\includegraphics[width=0.243\columnwidth]{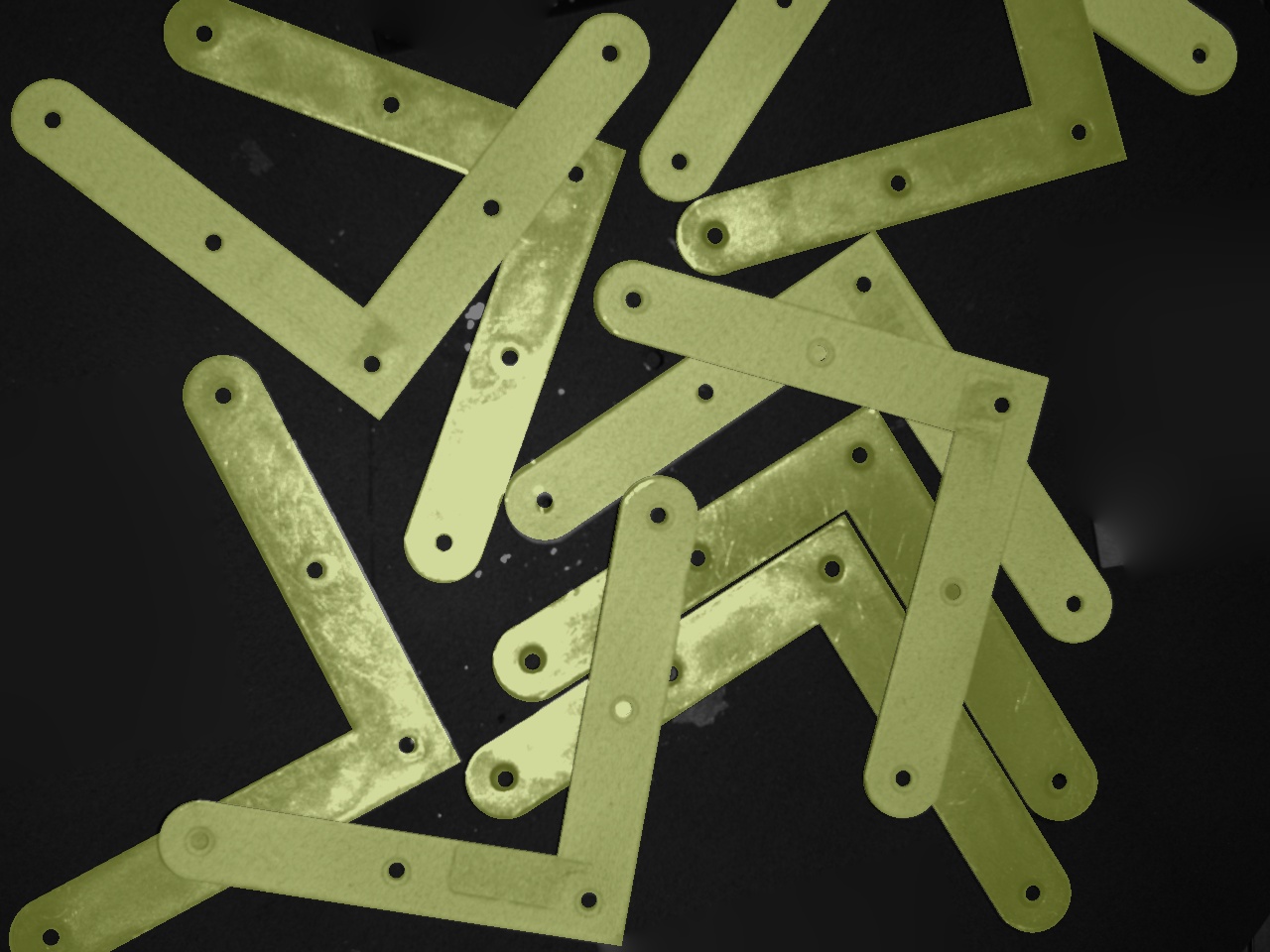} \vspace{0.1ex} \\
\end{tabular}

\begin{tabular}{ @{}c@{ } @{}c@{ } @{}c@{ } @{}c@{ } }
HB*~\cite{kaskman2019homebreweddb} & YCB-V*~\cite{xiang2017posecnn} & RU-APC~\cite{rennie2016dataset} & IC-BIN*~\cite{doumanoglou2016recovering} \vspace{0.5ex} \\
\includegraphics[width=0.243\columnwidth]{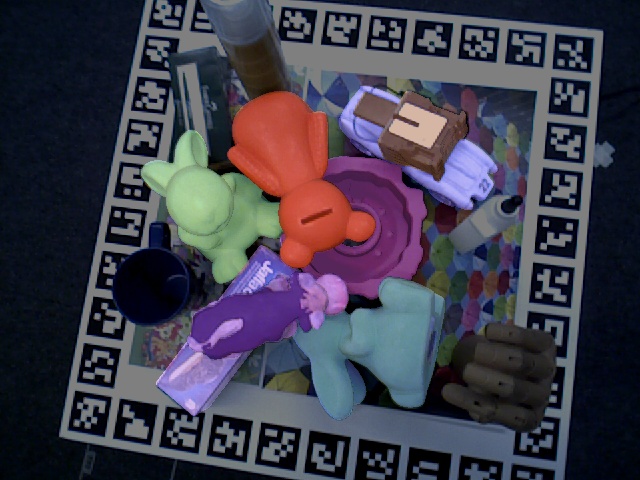} &
\includegraphics[width=0.243\columnwidth]{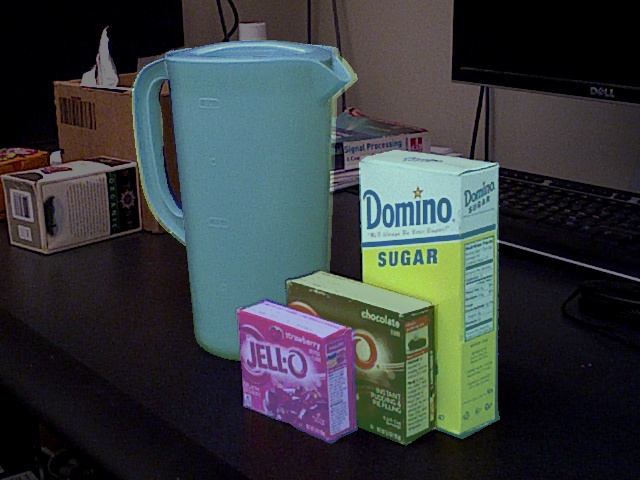} &
\includegraphics[width=0.243\columnwidth]{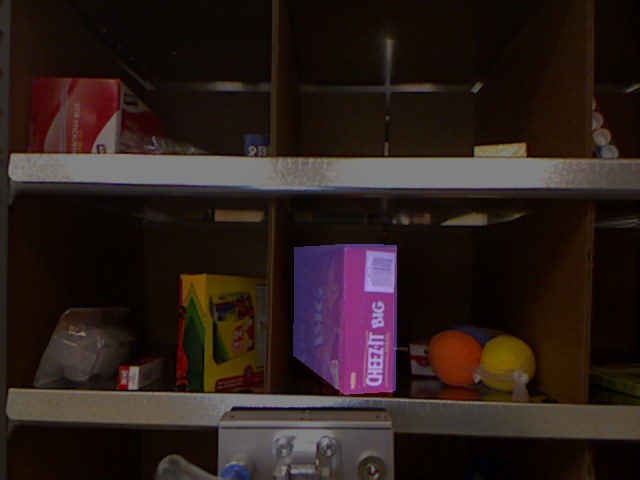} &
\includegraphics[width=0.243\columnwidth]{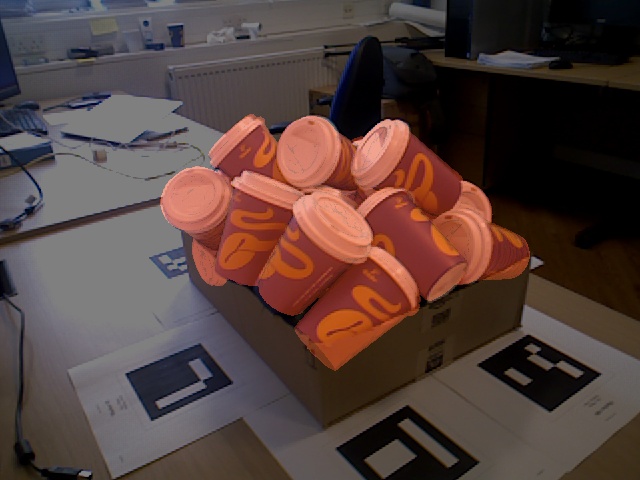} \vspace{0.1ex} \\
\end{tabular}

\begin{tabular}{ @{}c@{ } @{}c@{ } @{}c@{ } @{}c@{ }}
IC-MI~\cite{tejani2014latent} & TUD-L*~\cite{hodan2018bop} & TYO-L~\cite{hodan2018bop} & HOPE~\cite{tyree2022hope}\vspace{0.5ex} \\
\includegraphics[width=0.243\columnwidth]{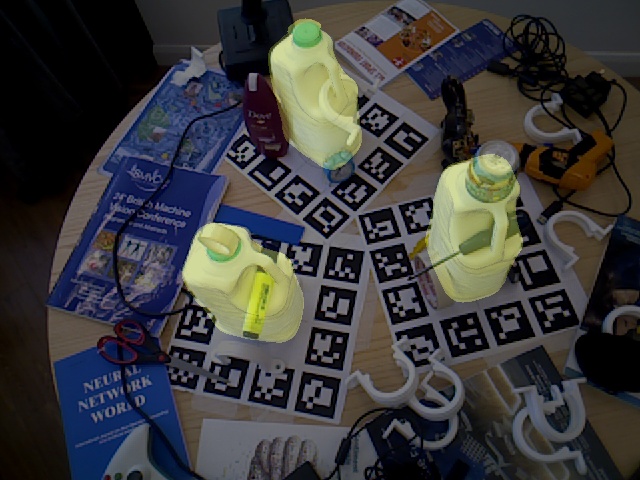} &
\includegraphics[width=0.243\columnwidth]{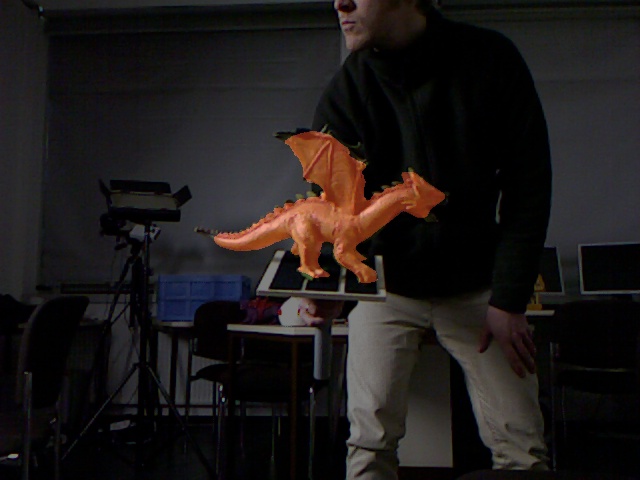} &
\includegraphics[width=0.243\columnwidth]{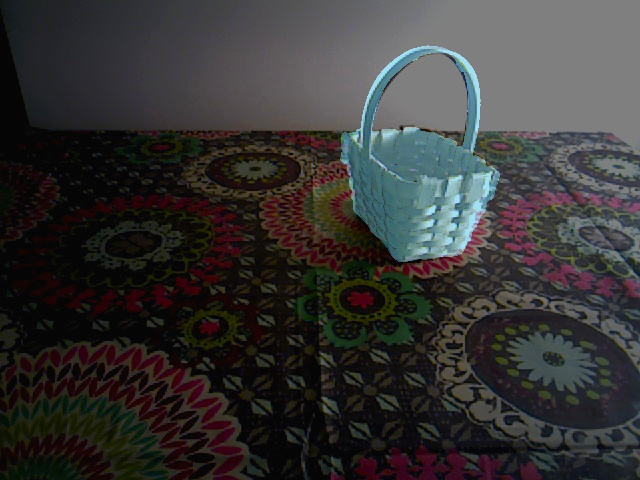} &
\includegraphics[width=0.243\columnwidth]{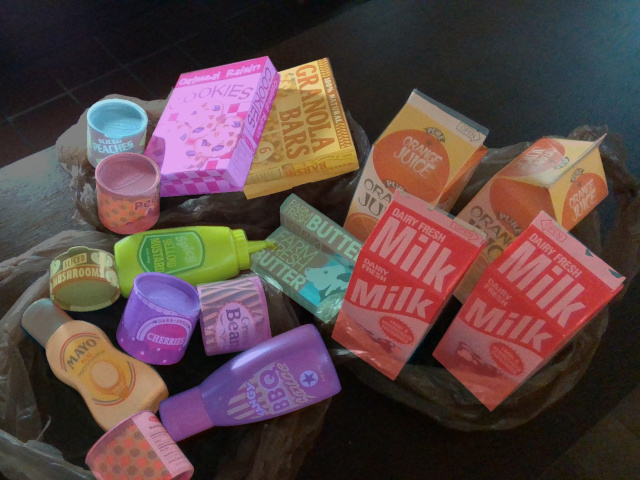} \\
\end{tabular}

\endgroup

\caption{\label{fig:dataset_overview}
\textbf{An overview of the BOP datasets.} The seven core datasets are marked with a star. Shown are RGB channels of sample test images which were darkened and overlaid with colored 3D object models in the ground-truth 6D poses.
\vspace{4ex}
}

\begingroup
\setlength\tabcolsep{1.0pt}
\footnotesize
\begin{tabularx}{\columnwidth}{ l r R R R R R R R }
	\toprule
	&
	&
	\multicolumn{2}{c}{Train. im.} &
	\multicolumn{1}{c}{Val im.} &
	\multicolumn{2}{c}{Test im.} &
	\multicolumn{2}{c}{Test inst.} \\
	\cmidrule(l{2pt}r{2pt}){3-4} \cmidrule(l{2pt}r{2pt}){5-5} \cmidrule(l{2pt}r{2pt}){6-7} \cmidrule(l{2pt}r{2pt}){8-9}
	\multicolumn{1}{l}{Dataset} &
	\multicolumn{1}{c}{Obj.} &
	\multicolumn{1}{c}{Real} &
	\multicolumn{1}{c}{PBR} &
	\multicolumn{1}{c}{Real} &
	\multicolumn{1}{c}{All} &
	\multicolumn{1}{c}{Used} &
	\multicolumn{1}{c}{All} &
	\multicolumn{1}{c}{Used} \\
	\midrule

	LM-O \cite{brachmann2014learning} & 8 & -- & 50K & -- & 1214 & 200 & 9038 & 1445 \\
	T-LESS \cite{hodan2017tless} &  30 & 37584 & 50K & -- & 10080 & 1000 & 67308 & 6423 \\
	ITODD \cite{drost2017introducing} &  28 & -- & 50K & 54 & 721 & 721 & 3041 & 3041 \\
	HB \cite{kaskman2019homebreweddb} &  33 & -- & 50K & 4420 & 13000 & 300 & 67542 & 1630 \\
	YCB-V \cite{xiang2017posecnn} &  21 & 113198 & 50K & -- & 20738 & 900 & 98547 & 4123 \\
	TUD-L \cite{hodan2018bop} &  3 & 38288 & 50K & -- & 23914 & 600 & 23914 & 600 \\
	IC-BIN \cite{doumanoglou2016recovering} & 2 & -- & 50K & -- & 177 & 150 & 2176 & 1786 \\
	\midrule
	LM \cite{hinterstoisser2012accv} & 15 & -- & 50K & -- & 18273 & 3000 & 18273 & 3000 \\
	RU-APC \cite{rennie2016dataset} & 14 & -- & -- & -- & 5964 & 1380 & 5964 & 1380\\
	IC-MI \cite{tejani2014latent} & 6 & -- & -- & -- & 2067 & 300 & 5318 & 800 \\
	TYO-L \cite{hodan2018bop} & 21 & -- & -- & -- & 1670 & 1670 & 1670 & 1670 \\
        HOPE \cite{tyree2022hope} & 28 & -- & -- & 50 & 188 & 188 & 3472 & 2898 \\
	\bottomrule
\end{tabularx}

\endgroup

\captionof{table}{\label{tab:dataset_params} \textbf{Parameters of the BOP datasets.} The core datasets are listed in the upper part.
PBR training images rendered by BlenderProc~\cite{denninger2019blenderproc,denninger2020blenderproc} are provided for all core datasets.
If a dataset includes both validation and test images, ground-truth annotations are public only for the validation images. All test images are real.
Column ``Test inst./All'' shows the number of annotated object instances for which at least $10\%$ of the projected surface area is visible in the test image. Columns ``Used'' show the number of used test images and object instances. %
\vspace{4ex}
}

\begingroup
  \includegraphics[width=1.0\linewidth]{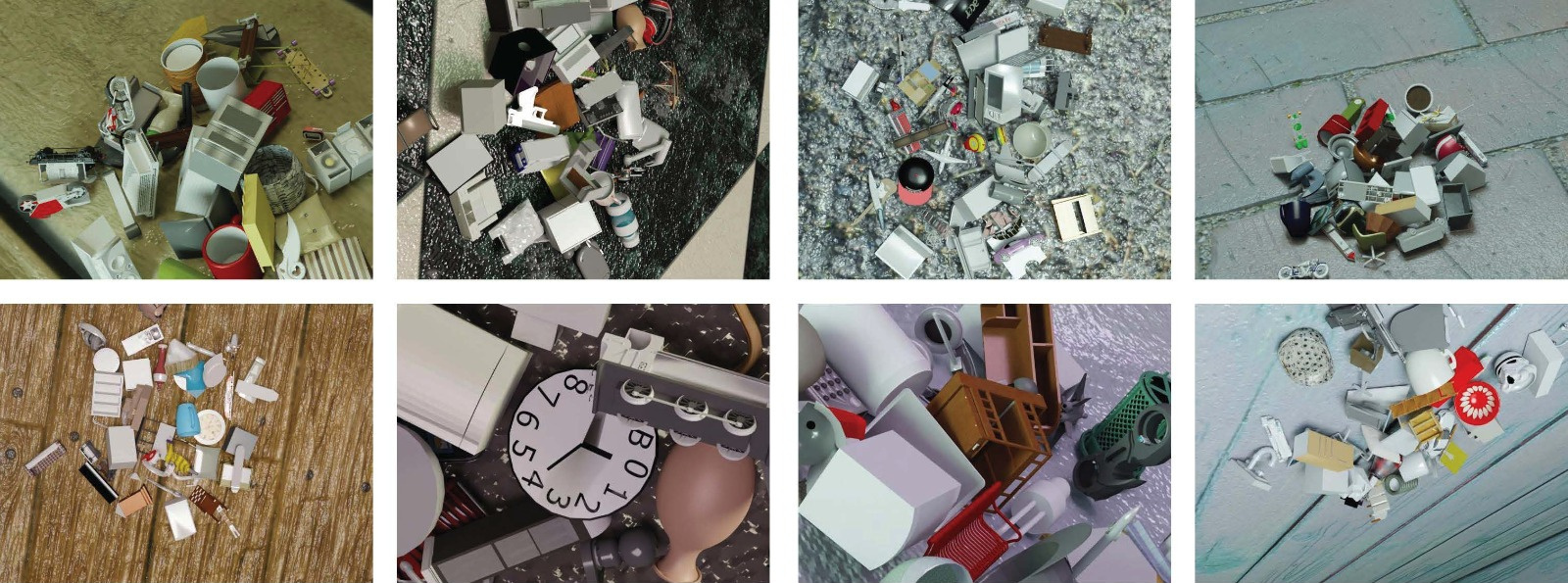}
\endgroup
\caption{\textbf{Example training images from the MegaPose dataset~\cite{megapose}.} This dataset includes 2M images showing annotated instances of more than 50K diverse objects and is meant for training methods for tasks on unseen objects (Tasks 4--6). The objects are not present in any other BOP dataset and their 3D models are available.}\label{fig:megapose_dataset}

\vspace{-4ex}
\end{center}
\end{figure}

\section{Datasets} \label{sec:datasets}

\subsection{Core datasets}

BOP currently includes twelve datasets in a unified format. Sample test images are in 
Fig.~\ref{fig:dataset_overview}
and dataset parameters in
Tab.~\ref{tab:dataset_params}.
Seven from the twelve were selected as core datasets: LM-O, T-LESS, ITODD, HB, YCB-V, TUD-L, and IC-BIN. Since 2019, methods must be evaluated on all of these core datasets to be considered for the main challenge awards (Sec.~\ref{sec:awards}).

Each dataset includes 3D object models and training and test RGB-D images annotated with ground-truth 6D object poses. The object models are provided in the form of 3D meshes (in most cases with a color texture) which were created manually or using KinectFusion-like systems for 3D reconstruction~\cite{newcombe2011kinectfusion}. While all test images are real, training images may be real and/or synthetic.
The seven core datasets include a total of 350K photorealistic PBR (physically-based rendered) training images generated and automatically annotated with BlenderProc~\cite{denninger2019blenderproc,denninger2020blenderproc,denninger2023blenderproc2}.
Example images, a description of the generation process and an analysis of the importance of PBR training images are in Sec. 3.2 and 4.3 of the 2020 challenge paper~\cite{hodan2020bop}. Datasets T-LESS, TUD-L and YCB-V include also real training images, and most datasets additionally include training images obtained by OpenGL rendering of the 3D object models on a black background.
Test images were captured in scenes with graded complexity, often with clutter and occlusion. Datasets HB and ITODD include also real validation images -- in this case, the ground-truth poses are publicly available only for the validation and not for the test images. The datasets can be downloaded from the BOP website
and more details can be found in Chapter 7 of~\cite{hodan2021phd}.

\begin{figure*}[t!]
\begin{center}
    \begingroup
    \setlength{\tabcolsep}{0.0pt}
    \renewcommand{\arraystretch}{1.0}
    \newlength{\plotwidth}
    \setlength{\plotwidth}{0.140\linewidth}
    \begin{tabular}{c c c c c c}
    \small{Ground-truth} & \small{GPose~\cite{gpose2023}} & \small{GenFlow~\cite{genflow}} & 
    \small{Ground-truth} & \small{GPose~\cite{gpose2023}} & \small{GenFlow~\cite{genflow}} \\
 
        \frame{\includegraphics[trim={50mm 10mm 0mm 0mm}, clip,height=\plotwidth]{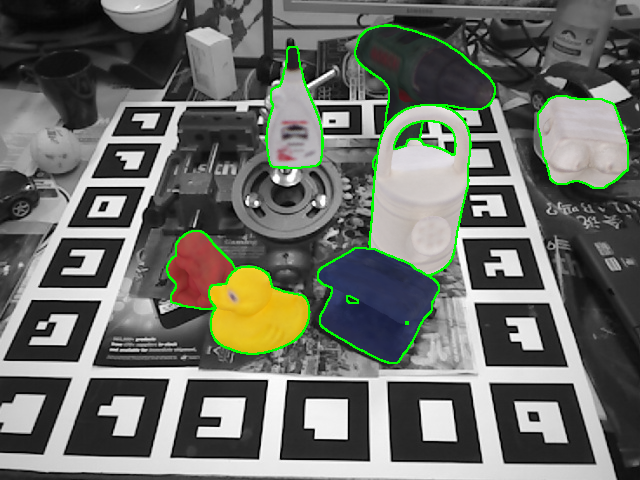}} & \hspace{0.1ex}
        \frame{\includegraphics[trim={50mm 10mm 0mm 0mm}, clip,height=\plotwidth]{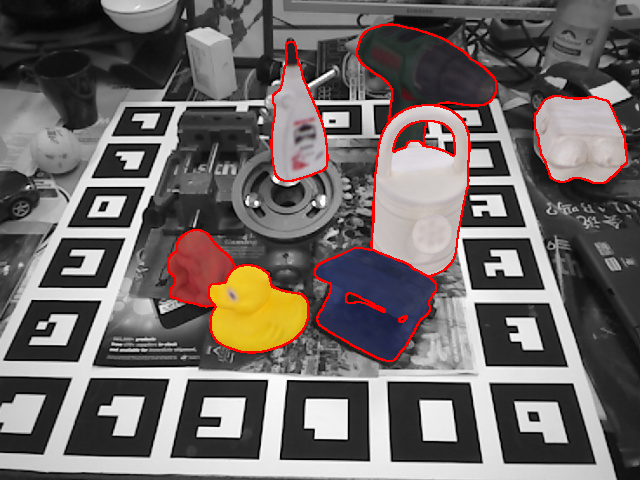}} & \hspace{0.1ex}
        \frame{\includegraphics[trim={50mm 10mm 0mm 0mm}, clip,height=\plotwidth]{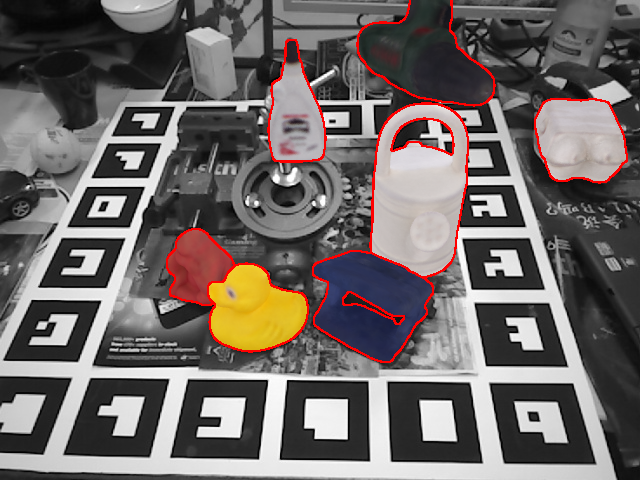}} & \hspace{2ex}
        \frame{\includegraphics[trim={10mm 0mm 25mm 0mm}, clip,height=\plotwidth]{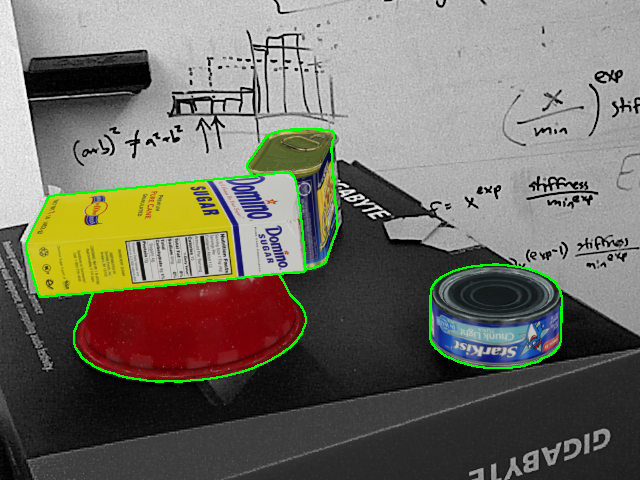}} & \hspace{0.1ex}
        \frame{\includegraphics[trim={10mm 0mm 25mm 0mm}, clip,height=\plotwidth]{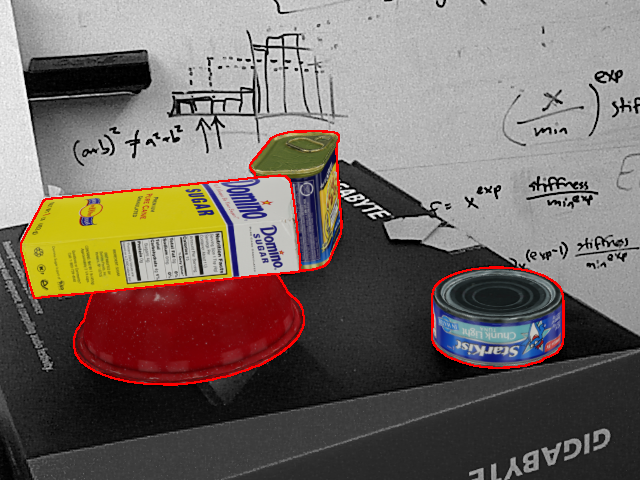}} & \hspace{0.1ex}
        \frame{\includegraphics[trim={10mm 0mm 25mm 0mm}, clip,height=\plotwidth]{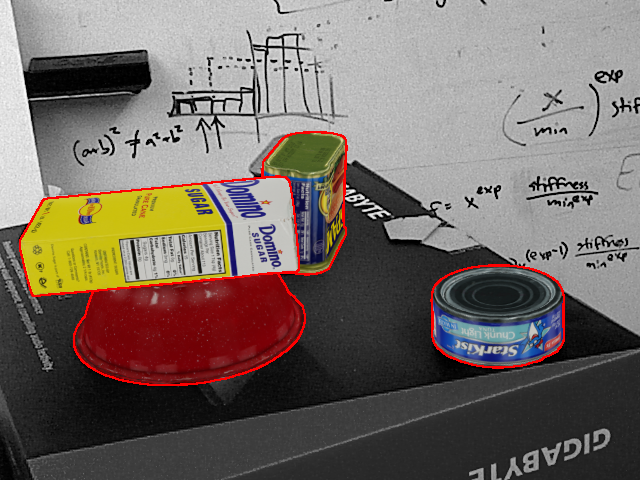}} \\
        
        \frame{\includegraphics[trim={50mm 10mm 0mm 0mm}, clip,height=\plotwidth]{img/bop23_qualitative/lmo/000002_000338_rgb.png}} & \hspace{0.1ex}
        \frame{\includegraphics[trim={50mm 10mm 0mm 0mm}, clip,height=\plotwidth]{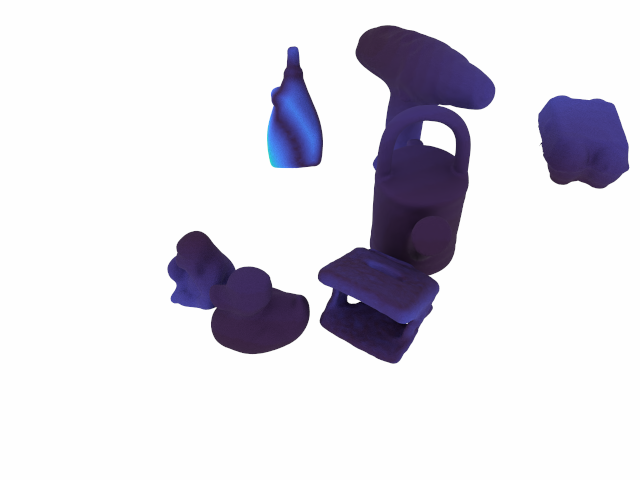}} & \hspace{0.1ex}
        \frame{\includegraphics[trim={50mm 10mm 0mm 0mm}, clip,height=\plotwidth]{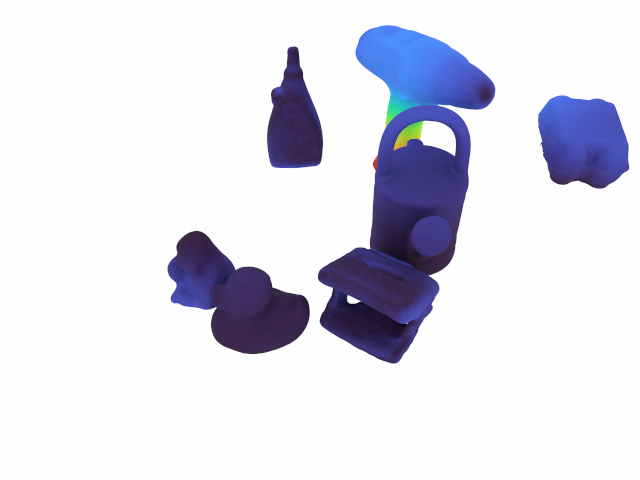}} & \hspace{2ex}
        \frame{\includegraphics[trim={10mm 0mm 25mm 0mm}, clip,height=\plotwidth]{img/bop23_qualitative/ycbv/000049_002089_rgb.png}} & \hspace{0.1ex}
        \frame{\includegraphics[trim={10mm 0mm 25mm 0mm}, clip,height=\plotwidth]{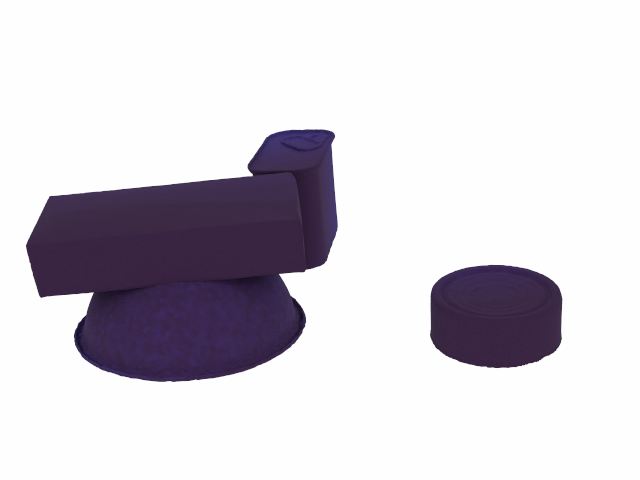}} & \hspace{0.1ex}
        \frame{\includegraphics[trim={10mm 0mm 25mm 0mm}, clip,height=\plotwidth]{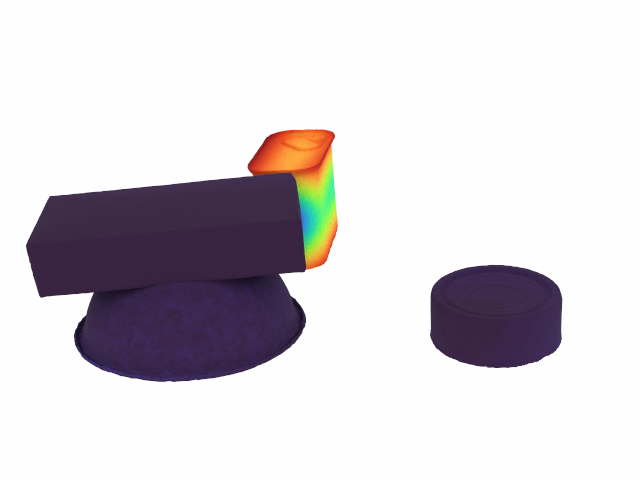}} \\
    \end{tabular}
    \endgroup
    \caption{
    \textbf{Qualitative comparison of the state-of-the-art methods for 6D localization of seen (GPose) and unseen objects (GenFlow)} on sample images from LM-O~\cite{brachmann2014learning} and YCB-V~\cite{xiang2017posecnn}. The bottom row shows the depth error map of each estimated pose \wrt the ground-truth pose. The map shows the distance between each 3D point in the ground-truth depth map and its position in the estimated pose (darker red indicates higher error: 0~cm \includegraphics[height=1.8mm]{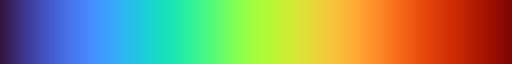} 10~cm). While GenFlow demonstrates strong performance on unseen objects, it tends to fail on challenging cases with heavy object occlusion (\eg, the drill in the sample LM-O image or the meat can in the YCB-V image).}
    \label{fig:cover_results}
\end{center}
\vspace{-1.0em}
\end{figure*}

\subsection{Training dataset for tasks on unseen objects}

In 2023, as a training dataset for Tasks 4--6 (Sec.~\ref{sec:methodology}), we provided over 2M images in the BOP format showing more than 50K diverse objects (Fig.~\ref{fig:megapose_dataset}). The images were originally synthesized for MegaPose~\cite{megapose} using BlenderProc~\cite{denninger2019blenderproc,denninger2020blenderproc,denninger2023blenderproc2}. The objects are from the Google Scanned Objects~\cite{downs2022google} and ShapeNetCore~\cite{chang2015shapenet} datasets. Note that symmetry transformations are not available for these objects, but could be identified as described in Sec.~2.3 of~\cite{hodan2020bop}.

\section{Results and discussion} \label{sec:evaluation}
This section presents results of the BOP Challenge 2023, compares them with results from earlier challenge editions, and summarizes the main messages for our field.
In total, 65 methods were fully evaluated (on all seven core datasets) on Task~1; 9 methods on Task~2; 11 methods on Task~3; 14 methods on Task~4; 3 methods on Task~5 and 4 methods on Task~6. Note that some of the results on Tasks 1--3 are from previous editions of the challenge.

\subsection{Experimental setup}
Participants of the 2023 challenge were submitting results
to the online evaluation system at \texttt{\href{http://bop.felk.cvut.cz/}{bop.felk.cvut.cz}} from June 7, 2023 until the deadline on September 28, 2023. The evaluation scripts are publicly available in the BOP toolkit\footnote{\texttt{\href{https://github.com/thodan/bop_toolkit}{github.com/thodan/bop\_toolkit}}}.

A method had to use a fixed set of hyper-parameters across all objects and datasets. 
For the tasks on seen objects (Tasks 1--3), a method could use the provided 3D object models and training images as well as render extra unlimited training images.
For the tasks on unseen objects (Tasks 4--6), a method had to onboard new objects from their 3D models in a limited onboarding stage of 5 minutes on a PC with a single GPU. The method could render images of the 3D models or use a subset of the BlenderProc images originally provided for BOP 2020~\cite{hodan2020bop} -- the method could use as many images from this set as could be rendered within the limited onboarding time (rendering and any additional processing had to fit within 5 minutes, considering that rendering of one BlenderProc image takes 2 seconds).

Not a single pixel of test images may have been used for training and onboarding, nor the individual ground-truth annotations that are publicly available for test images of some datasets.
Ranges of the azimuth and elevation camera angles, and a range of the camera-object distances determined by the ground-truth poses from test images are the only information about the test set that may have been used during training and onboarding.
Only subsets of test images were used
(see Tab.~\ref{tab:dataset_params})
to remove redundancies and speed up the evaluation, and only object instances for which at least $10\%$ of the projected surface area is visible were considered in the evaluation.

\setlength{\tabcolsep}{2pt}
\newcounter{mycounter}
\begin{table*}[t!]
    \tiny
    \centering
    \begin{tabularx}{\linewidth}{rllllllll lYYYYYYYYY}
        \toprule
        \# & Method & Year & Type & DNN per & Det./seg. & Refinement & Train im. & ...type & Test im. & LM-O & T-LESS & TUD-L & IC-BIN & ITODD & HB & YCB-V & \mbox{$\text{AR}_C$} & Time \\ 
        \midrule
        1 & GPose2023 \cite{Wang_2021_GDRN,liu2022gdrnpp_bop} & \cellcolor{ccol!100}2023 & DNN & Object & Custom & \cellcolor{ccol!100}${\tiny\sim}$Coord-guided & RGB-D & PBR+real & RGB-D & \cellcolor{arcol!79.4}79.4 & \cellcolor{arcol!91.4}91.4 & \cellcolor{arcol!96.4}96.4 & \cellcolor{arcol!73.7}73.7 & \cellcolor{arcol!70.4}70.4 & \cellcolor{arcol!95.0}95.0 & \cellcolor{arcol!92.8}92.8 & \cellcolor{avgcol!85.6}85.6 & \cellcolor{timecol!26.70}$\phantom{0}$2.67 \\ 
        
        2 & GPose2023-OfficialDet \cite{Wang_2021_GDRN,liu2022gdrnpp_bop} & \cellcolor{ccol!100}2023 & DNN & Object & \cellcolor{ccol!100}Default GDRNPPDet & \cellcolor{ccol!100}${\tiny\sim}$Coord-guided & RGB-D & PBR+real & RGB-D & \cellcolor{arcol!80.5}80.5 & \cellcolor{arcol!89.5}89.5 & \cellcolor{arcol!96.6}96.6 & \cellcolor{arcol!73.4}73.4 & \cellcolor{arcol!68.7}68.7 & \cellcolor{arcol!94.4}94.4 & \cellcolor{arcol!92.9}92.9 & \cellcolor{avgcol!85.1}85.1 & \cellcolor{timecol!45.75}$\phantom{0}$4.57 \\ 
        
        3 & GPose2023-PBR \cite{Wang_2021_GDRN,liu2022gdrnpp_bop} & \cellcolor{ccol!100}2023 & DNN & Object & Custom & \cellcolor{ccol!100}${\tiny\sim}$Coord-guided & RGB-D & PBR+real & RGB-D & \cellcolor{arcol!79.4}79.4 & \cellcolor{arcol!89.0}89.0 & \cellcolor{arcol!93.1}93.1 & \cellcolor{arcol!73.7}73.7 & \cellcolor{arcol!70.4}70.4 & \cellcolor{arcol!95.0}95.0 & \cellcolor{arcol!90.1}90.1 & \cellcolor{avgcol!84.4}84.4 & \cellcolor{timecol!26.86}$\phantom{0}$2.86 \\ 
        
        4 & GDRNPP-PBRReal-RGBD-MModel \cite{Wang_2021_GDRN,liu2022gdrnpp_bop} & 2022 & DNN & Object & YOLOX & \cellcolor{ccol!100}${\tiny\sim}$CIR & RGB-D & PBR+real & RGB-D & \cellcolor{arcol!77.5}77.5	&\cellcolor{arcol!87.4}87.4	&\cellcolor{arcol!96.6}96.6	&\cellcolor{arcol!72.2}72.2&	\cellcolor{arcol!67.9}67.9&	\cellcolor{arcol!92.6}92.6&	\cellcolor{arcol!92.1}92.1& \cellcolor{avgcol!83.7}83.7 & \cellcolor{timecol!62.63}$\phantom{0}$6.26 \\ 
        
        5 & GDRNPP-PBR-RGBD-MModel \cite{Wang_2021_GDRN,liu2022gdrnpp_bop} & 2022 & DNN & Object & YOLOX & \cellcolor{ccol!100}${\tiny\sim}$CIR & RGB-D & \cellcolor{ccol!100}PBR & RGB-D & \cellcolor{arcol!77.5}77.5	&\cellcolor{arcol!85.2}85.2	&\cellcolor{arcol!92.9}92.9	&\cellcolor{arcol!72.2}72.2&	\cellcolor{arcol!67.9}67.9&	\cellcolor{arcol!92.6}92.6&	\cellcolor{arcol!90.6}90.6& \cellcolor{avgcol!82.7}82.7 & \cellcolor{timecol!62.64}$\phantom{0}$6.26 \\ 
        
        6 & ZebraPoseSAT-EffnetB4-refined~\cite{su2022zebrapose} & \cellcolor{ccol!100}2023 & DNN & Object & \cellcolor{ccol!100}Default GDRNPPDet & \cellcolor{ccol!100}\cellcolor{ccol!100}${\tiny\sim}$CIR & RGB-D & PBR+real & RGB-D  & \cellcolor{arcol!78.0}78.0 & \cellcolor{arcol!86.2}86.2 & \cellcolor{arcol!95.6}95.6 & \cellcolor{arcol!65.4}65.4 & \cellcolor{arcol!61.8}61.8 & \cellcolor{arcol!92.1}92.1 & \cellcolor{arcol!89.9}89.9 & \cellcolor{avgcol!81.3}81.3 & \cellcolor{timecol!25.77}$\phantom{0}$2.57 \\
        
        7 & GDRNPP-PBRReal-RGBD-MModel-Fast \cite{Wang_2021_GDRN,liu2022gdrnpp_bop} & 2022 & DNN & Object & YOLOX & \cellcolor{ccol!100}Depth adjust. & \cellcolor{ccol!100}RGB & PBR+real & RGB-D & \cellcolor{arcol!79.2}79.2	&\cellcolor{arcol!87.2}87.2	&\cellcolor{arcol!93.6}93.6	&\cellcolor{arcol!70.2}70.2&	\cellcolor{arcol!58.8}58.8&	\cellcolor{arcol!90.9}90.9&	\cellcolor{arcol!83.4}83.4& \cellcolor{avgcol!80.5}80.5 & \cellcolor{timecol!02.28}$\phantom{0}$0.23 \\ 
        
        8 & OfficialDet-PFA-Mixpbr-RGB-D~\cite{hu2022perspective} & \cellcolor{ccol!100}2023 & DNN & \cellcolor{ccol!100}Dataset & \cellcolor{ccol!100}Default GDRNPPDet & \cellcolor{ccol!100}PFA & \cellcolor{ccol!100}RGB & PBR+real & RGB-D  & \cellcolor{arcol!79.2}79.2 & \cellcolor{arcol!84.9}84.9 & \cellcolor{arcol!96.3}96.3 & \cellcolor{arcol!70.6}70.6 & \cellcolor{arcol!52.6}52.6 & \cellcolor{arcol!86.7}86.7 & \cellcolor{arcol!89.9}89.9 & \cellcolor{avgcol!80.0}80.0 & \cellcolor{timecol!11.93}$\phantom{0}$1.19 \\
        
        9 & GDRNPP-PBRReal-RGBD-MModel-Offi. \cite{Wang_2021_GDRN,liu2022gdrnpp_bop} & 2022 & DNN & Object & \cellcolor{ccol!100}Default (synt+real) & \cellcolor{ccol!100}${\tiny\sim}$CIR & RGB-D & PBR+real & RGB-D & \cellcolor{arcol!75.8}75.8	&\cellcolor{arcol!82.4}82.4	&\cellcolor{arcol!96.6}96.6	&\cellcolor{arcol!70.8}70.8&	\cellcolor{arcol!54.3}54.3&	\cellcolor{arcol!89.0}89.0&	\cellcolor{arcol!89.6}89.6& \cellcolor{avgcol!79.8}79.8 & \cellcolor{timecol!64.06}$\phantom{0}$6.41 \\
        10 & GDRNPPDet-PBRReal+GenFlow-MultiHypo~\cite{liu2022gdrnpp_bop} & \cellcolor{ccol!100}2023 & DNN & \cellcolor{ccol!100}Dataset & \cellcolor{ccol!100}Default (synt+real) & \cellcolor{ccol!100}Recurrent Flow & RGB-D & PBR+real & RGB-D  & \cellcolor{arcol!74.4}74.4 & \cellcolor{arcol!78.0}78.0 & \cellcolor{arcol!92.4}92.4 & \cellcolor{arcol!65.1}65.1 & \cellcolor{arcol!64.7}64.7 & \cellcolor{arcol!91.6}91.6 & \cellcolor{arcol!88.4}88.4 & 
        \cellcolor{avgcol!79.2}79.2 & \cellcolor{timecol!100}36.01\\
        
        11 & Extended\_FCOS+PFA-MixPBR-RGBD~\cite{hu2022perspective} & 2022 & DNN & \cellcolor{ccol!100}Dataset & Extended FCOS & \cellcolor{ccol!100}PFA & \cellcolor{ccol!100}RGB & PBR+real & RGB-D & \cellcolor{arcol!79.7}79.7	&\cellcolor{arcol!85.0}85.0	&\cellcolor{arcol!96.0}96.0	&\cellcolor{arcol!67.6}67.6&	\cellcolor{arcol!46.9}46.9&	\cellcolor{arcol!86.9}86.9&	\cellcolor{arcol!88.8}88.8& \cellcolor{avgcol!78.7}78.7 & \cellcolor{timecol!23.17}$\phantom{0}$2.32 \\ 
        
        12  & Extended\_FCOS+PFA-MixPBR-RGBD-Fast~\cite{hu2022perspective} & 2022 & DNN & \cellcolor{ccol!100}Dataset & Extended FCOS & \cellcolor{ccol!100}PFA & \cellcolor{ccol!100}RGB & PBR+real & RGB-D  & \cellcolor{arcol!79.2}79.2	&\cellcolor{arcol!77.9}77.9	&\cellcolor{arcol!95.8}95.8	&\cellcolor{arcol!67.1}67.1&	\cellcolor{arcol!46.0}46.0&	\cellcolor{arcol!86.0}86.0&	\cellcolor{arcol!88.0}88.0& \cellcolor{avgcol!77.1}77.1 & \cellcolor{timecol!06.39}$\phantom{0}$0.64 \\ 
        
       13 & RCVPose3D-SingleModel-VIVO-PBR~\cite{wu2022keypoint} & 2022 & DNN & \cellcolor{ccol!100}Dataset & RCVPose3D & \cellcolor{ccol!100}ICP & RGB-D & PBR+real & RGB-D  & \cellcolor{arcol!72.9}72.9	&\cellcolor{arcol!70.8}70.8	&\cellcolor{arcol!96.6}96.6	&\cellcolor{arcol!73.3}73.3&	\cellcolor{arcol!53.6}53.6&	\cellcolor{arcol!86.3}86.3&	\cellcolor{arcol!84.3}84.3& \cellcolor{avgcol!76.8}76.8 & \cellcolor{timecol!13.36}$\phantom{0}$1.34 \\ 
       
        14 & ZebraPoseSAT-EffnetB4+ICP(DefaultDet)~\cite{su2022zebrapose} & 2022 & DNN & Object & \cellcolor{ccol!100}Default (synt+real) & \cellcolor{ccol!100}ICP & \cellcolor{ccol!100}RGB & PBR+real & RGB-D & \cellcolor{arcol!75.2}75.2	&\cellcolor{arcol!72.7}72.7	&\cellcolor{arcol!94.8}94.8	&\cellcolor{arcol!65.2}65.2&	\cellcolor{arcol!52.7}52.7&	\cellcolor{arcol!88.3}88.3&	\cellcolor{arcol!86.6}86.6& \cellcolor{avgcol!76.5}76.5 & \cellcolor{timecol!05.00}$\phantom{0}$0.50 \\
        
        15 & Extended\_FCOS+PFA-PBR-RGBD~\cite{hu2022perspective} & 2022 & DNN & \cellcolor{ccol!100}Dataset & Extended FCOS & \cellcolor{ccol!100}PFA & \cellcolor{ccol!100}RGB & \cellcolor{ccol!100}PBR & RGB-D & \cellcolor{arcol!79.7}79.7	&\cellcolor{arcol!80.2}80.2	&\cellcolor{arcol!89.3}89.3	&\cellcolor{arcol!67.6}67.6&	\cellcolor{arcol!46.9}46.9&	\cellcolor{arcol!86.9}86.9&	\cellcolor{arcol!82.6}82.6& \cellcolor{avgcol!76.2}76.2 & \cellcolor{timecol!26.31}$\phantom{0}$2.63 \\ 
        
        16 & SurfEmb-PBR-RGBD~\cite{haugaard2022surfemb} & 2022 & DNN & \cellcolor{ccol!100}Dataset & \cellcolor{ccol!100}Default (PBR) & \cellcolor{ccol!100}Custom & RGB-D & \cellcolor{ccol!100}PBR & RGB-D & \cellcolor{arcol!76.0}76.0	&\cellcolor{arcol!82.8}82.8	&\cellcolor{arcol!85.4}85.4	&\cellcolor{arcol!65.9}65.9&	\cellcolor{arcol!53.8}53.8&	\cellcolor{arcol!86.6}86.6&	\cellcolor{arcol!79.9}79.9& \cellcolor{avgcol!75.8}75.8 & \cellcolor{timecol!90.48}$\phantom{0}$9.05 \\
        
        17 & ZebraPoseSAT-EffnetB4~\cite{su2022zebrapose}  & \cellcolor{ccol!100}2023 & DNN & Object & \cellcolor{ccol!100}Default GDRNPPDet & -- & \cellcolor{ccol!100}RGB & \cellcolor{ccol!100}PBR+real & \cellcolor{ccol!100}RGB  & \cellcolor{arcol!72.9}72.9 & \cellcolor{arcol!82.1}82.1 & \cellcolor{arcol!85.0}85.0 & \cellcolor{arcol!59.2}59.2 & \cellcolor{arcol!50.4}50.4 & \cellcolor{arcol!92.2}92.2 & \cellcolor{arcol!82.8}82.8 &
        \cellcolor{avgcol!74.9}74.9 & 
        \cellcolor{timecol!25.0}$\phantom{0}$2.50\\
        
        18 & GDRNPP-PBRReal-RGBD-SModel \cite{Wang_2021_GDRN,liu2022gdrnpp_bop} & 2022 & DNN & \cellcolor{ccol!100}Dataset & YOLOX & \cellcolor{ccol!100}Depth adjust. & \cellcolor{ccol!100}RGB & PBR+real & RGB-D  & \cellcolor{arcol!75.7}75.7	&\cellcolor{arcol!85.6}85.6	&\cellcolor{arcol!90.6}90.6	&\cellcolor{arcol!68.0}68.0&	\cellcolor{arcol!35.6}35.6&	\cellcolor{arcol!86.4}86.4&	\cellcolor{arcol!81.7}81.7& \cellcolor{avgcol!74.8}74.8 & \cellcolor{timecol!05.56}$\phantom{0}$0.56 \\
        
        19 & Megapose-GDRNPPDet-PBRReal+Multi~\cite{megapose,liu2022gdrnpp_bop} & \cellcolor{ccol!100}2023 & DNN & \cellcolor{ccol!100}Dataset & \cellcolor{ccol!100}Default GDRNPPDet & \cellcolor{ccol!100}Teaser++ & \cellcolor{ccol!100}RGB & PBR & RGB-D &  \cellcolor{arcol!70.4}70.4 & \cellcolor{arcol!71.8}71.8 & \cellcolor{arcol!91.6}91.6 & \cellcolor{arcol!59.2}59.2 & \cellcolor{arcol!55.3}55.3 & \cellcolor{arcol!87.2}87.2 & \cellcolor{arcol!85.5}85.5 & 
        \cellcolor{avgcol!74.4}74.4 & \cellcolor{timecol!100}93.26\\
        20 & Coupled Iterative Refinement (CIR) \cite{lipson2022coupled} & 2022 & DNN & \cellcolor{ccol!100}Dataset & \cellcolor{ccol!100}Default (synt+real) & \cellcolor{ccol!100}CIR & RGB-D & PBR+real & RGB-D & \cellcolor{arcol!73.4}73.4	&\cellcolor{arcol!77.6}77.6	&\cellcolor{arcol!96.8}96.8	&\cellcolor{arcol!67.6}67.6&	\cellcolor{arcol!38.1}38.1&	\cellcolor{arcol!75.7}75.7&	\cellcolor{arcol!89.3}89.3& \cellcolor{avgcol!74.1}74.1& $\phantom{0}$-- \\ 
        
        21 & GPose2023-RGB~\cite{Wang_2021_GDRN,liu2022gdrnpp_bop}  & \cellcolor{ccol!100}2023 & DNN & Object & GDet2023 & \cellcolor{ccol!100}CIR & \cellcolor{ccol!100}RGB & PBR & \cellcolor{ccol!100}RGB &  \cellcolor{arcol!69.9}69.9 & \cellcolor{arcol!79.9}79.9 & \cellcolor{arcol!83.1}83.1 & \cellcolor{arcol!62.6}62.6 & \cellcolor{arcol!46.0}46.0 & \cellcolor{arcol!87.6}87.6 & \cellcolor{arcol!80.9}80.9 & 
        \cellcolor{avgcol!72.9}72.9 & \cellcolor{timecol!2.43}$\phantom{0}$0.24\\
        
        22 & GDRNPP-PBRReal-RGB-MModel\cite{Wang_2021_GDRN,liu2022gdrnpp_bop} & 2022 & DNN & Object & YOLOX & -- & \cellcolor{ccol!100}RGB & PBR+real & \cellcolor{ccol!100}RGB & \cellcolor{arcol!71.3}71.3	&\cellcolor{arcol!78.6}78.6	&\cellcolor{arcol!83.1}83.1	&\cellcolor{arcol!62.3}62.3&	\cellcolor{arcol!44.8}44.8&	\cellcolor{arcol!86.9}86.9&	\cellcolor{arcol!82.5}82.5& \cellcolor{avgcol!72.8}72.8 & \cellcolor{timecol!02.29}$\phantom{0}$0.23 \\ 
        
        23 & ZebraPoseSAT-EffnetB4~\cite{su2022zebrapose} & 2022 & DNN & Object & FCOS & -- & \cellcolor{ccol!100}RGB & PBR+real & \cellcolor{ccol!100}RGB & \cellcolor{arcol!72.1}72.1	&\cellcolor{arcol!80.6}80.6	&\cellcolor{arcol!85.0}85.0	&\cellcolor{arcol!54.5}54.5&	\cellcolor{arcol!41.0}41.0&	\cellcolor{arcol!88.2}88.2&	\cellcolor{arcol!83.0}83.0& \cellcolor{avgcol!72.0}72.0 & \cellcolor{timecol!02.50}$\phantom{0}$0.25 \\ 
        
        24 & ZebraPoseSAT-EffnetB4(DefaultDet)~\cite{su2022zebrapose} & 2022 & DNN & Object & \cellcolor{ccol!100}Default (synt+real) & -- & \cellcolor{ccol!100}RGB & PBR+real & \cellcolor{ccol!100}RGB & \cellcolor{arcol!70.7}70.7	&\cellcolor{arcol!76.8}76.8	&\cellcolor{arcol!84.9}84.9	&\cellcolor{arcol!59.7}59.7&	\cellcolor{arcol!41.7}41.7&	\cellcolor{arcol!88.7}88.7&	\cellcolor{arcol!81.6}81.6& \cellcolor{avgcol!72.0}72.0 & \cellcolor{timecol!02.50}$\phantom{0}$0.25 \\ 
        
        25 & ZebraPoseSAT-EffnetB4(PBR-DefaultDet)~\cite{su2022zebrapose} & \cellcolor{ccol!100}2023 & DNN & Object & \cellcolor{ccol!100} Default GDRNPPDet & -- & \cellcolor{ccol!100}RGB & PBR+real & \cellcolor{ccol!100}RGB &   \cellcolor{arcol!72.9}72.9 & \cellcolor{arcol!81.1}81.1 & \cellcolor{arcol!75.6}75.6 & \cellcolor{arcol!59.2}59.2 & \cellcolor{arcol!50.4}50.4 & \cellcolor{arcol!92.1}92.1 & \cellcolor{arcol!72.9}72.9 & 
        \cellcolor{avgcol!72.0}72.0 & \cellcolor{timecol!2.50}$\phantom{0}$0.25\\
        
          26 & ZebraPose-SAT~\cite{su2022zebrapose} & 2022 & DNN & Object & FCOS & -- & \cellcolor{ccol!100}RGB & PBR+real & \cellcolor{ccol!100}RGB & \cellcolor{arcol!72.1}72.1	&\cellcolor{arcol!78.7}78.7	&\cellcolor{arcol!86.1}86.1	&\cellcolor{arcol!54.9}54.9&	\cellcolor{arcol!37.9}37.9&	\cellcolor{arcol!84.7}84.7&	\cellcolor{arcol!82.8}82.8& \cellcolor{avgcol!71.0}71.0& $\phantom{0}$-- \\ 
          
         27 & Extended\_FCOS+PFA-MixPBR-RGB~\cite{hu2022perspective} & 2022 & DNN & \cellcolor{ccol!100}Dataset & Extended FCOS & \cellcolor{ccol!100}PFA & \cellcolor{ccol!100}RGB & PBR+real & \cellcolor{ccol!100}RGB  & \cellcolor{arcol!74.5}74.5	&\cellcolor{arcol!77.8}77.8	&\cellcolor{arcol!83.9}83.9	&\cellcolor{arcol!60.0}60.0&	\cellcolor{arcol!35.3}35.3&	\cellcolor{arcol!84.1}84.1&	\cellcolor{arcol!80.6}80.6& \cellcolor{avgcol!70.9}70.9 & \cellcolor{timecol!30.19}$\phantom{0}$3.02 \\ 
         
         28 & GDRNPP-PBR-RGB-MModel \cite{Wang_2021_GDRN,liu2022gdrnpp_bop} & 2022 & DNN & Object & YOLOX & -- & \cellcolor{ccol!100}RGB & \cellcolor{ccol!100}PBR & \cellcolor{ccol!100}RGB & \cellcolor{arcol!71.3}71.3	&\cellcolor{arcol!79.6}79.6	&\cellcolor{arcol!75.2}75.2	&\cellcolor{arcol!62.3}62.3&	\cellcolor{arcol!44.8}44.8&	\cellcolor{arcol!86.9}86.9&	\cellcolor{arcol!71.3}71.3& \cellcolor{avgcol!70.2}70.2 & \cellcolor{timecol!02.84}$\phantom{0}$0.28 \\ 
         
         29& GDRNPPDet-PBRReal+GenFlow-Multi~\cite{Wang_2021_GDRN,liu2022gdrnpp_bop} & \cellcolor{ccol!100}2023 & DNN & \cellcolor{ccol!100}Dataset & \cellcolor{ccol!100}Default GDRNPPDet & -- & \cellcolor{ccol!100}RGB & PBR+real & \cellcolor{ccol!100}RGB  &  \cellcolor{arcol!66.8}66.8 & \cellcolor{arcol!82.3}82.3 & \cellcolor{arcol!76.0}76.0 & \cellcolor{arcol!58.1}58.1 & \cellcolor{arcol!48.6}48.6 & \cellcolor{arcol!89.3}89.3 & \cellcolor{arcol!69.8}69.8 & 
        \cellcolor{avgcol!70.1}70.1 & \cellcolor{timecol!100}35.36 \\ 
         30 & CosyPose-ECCV20-SYNT+REAL-ICP~\cite{labbe2020cosypose} & 2020 & DNN & \cellcolor{ccol!100}Dataset & \cellcolor{ccol!100}Default (synt+real) & \cellcolor{ccol!100}DeepIM+ICP & \cellcolor{ccol!100}RGB & PBR+real & RGB-D  & \cellcolor{arcol!71.4}71.4	&\cellcolor{arcol!70.1}70.1	&\cellcolor{arcol!93.9}93.9	&\cellcolor{arcol!64.7}64.7&	\cellcolor{arcol!31.3}31.3&	\cellcolor{arcol!71.2}71.2&	\cellcolor{arcol!86.1}86.1& \cellcolor{avgcol!69.8}69.8 & \cellcolor{timecol!100}13.74 \\ 
          
        31 & MRPE-PBRReal-RGB-SModel & \cellcolor{ccol!100}2023 & DNN & \cellcolor{ccol!100}Dataset & \cellcolor{ccol!100}Default GDRNPPDet & \cellcolor{ccol!100}Render \& com. & \cellcolor{ccol!100}RGB & PBR+real & RGB-D &  \cellcolor{arcol!74.4}74.4 & \cellcolor{arcol!75.8}75.8 & \cellcolor{arcol!82.4}82.4 & \cellcolor{arcol!55.0}55.0 & \cellcolor{arcol!36.8}36.8 & \cellcolor{arcol!77.0}77.0 & \cellcolor{arcol!84.3}84.3 & 
        \cellcolor{avgcol!69.4}69.4 & \cellcolor{timecol!1.00}$\phantom{0}$0.10\\
        
       32 & GDRNPP-PBRReal-RGB-SModel & 2022 & DNN & \cellcolor{ccol!100}Dataset & YOLOX & \cellcolor{ccol!100}~CIR & \cellcolor{ccol!100}RGB & PBR+real & \cellcolor{ccol!100}RGB &   
        \cellcolor{arcol!68.6}68.6 & \cellcolor{arcol!77.6}77.6 & \cellcolor{arcol!82.7}82.7 & \cellcolor{arcol!61.7}61.7 & \cellcolor{arcol!26.0}26.0 & \cellcolor{arcol!80.9}80.9 & \cellcolor{arcol!76.8}76.8 & 
        \cellcolor{avgcol!67.8}67.8 & \cellcolor{timecol!4.66}$\phantom{0}$0.46\\
        33 & Megapose-GDRNPPDet-PBRReal+MultiHyp~\cite{megapose,liu2022gdrnpp_bop} & \cellcolor{ccol!100}2023 & DNN & \cellcolor{ccol!100}Dataset & \cellcolor{ccol!100}Default GDRNPPDet & -- & \cellcolor{ccol!100}RGB & \cellcolor{ccol!100}PBR & \cellcolor{ccol!100}RGB &   \cellcolor{arcol!64.8}64.8 & \cellcolor{arcol!78.1}78.1 & \cellcolor{arcol!74.1}74.1 & \cellcolor{arcol!56.9}56.9 & \cellcolor{arcol!42.2}42.2 & \cellcolor{arcol!86.3}86.3 & \cellcolor{arcol!70.2}70.2 & 
        \cellcolor{avgcol!67.5}67.5 & \cellcolor{timecol!100}36.28\\
        
        34 & ZebraPoseSAT-EffnetB4 (PBR\_Only)~\cite{su2022zebrapose} & 2022 & DNN & Object & FCOS & -- & \cellcolor{ccol!100}RGB & \cellcolor{ccol!100}PBR & \cellcolor{ccol!100}RGB  & \cellcolor{arcol!72.1}72.1	&\cellcolor{arcol!72.3}72.3	&\cellcolor{arcol!71.7}71.7	&\cellcolor{arcol!54.5}54.5&	\cellcolor{arcol!41.0}41.0&	\cellcolor{arcol!88.2}88.2&	\cellcolor{arcol!69.1}69.1& \cellcolor{avgcol!67.0}67.0& $\phantom{0}$-- \\
        35 & Extended\_FCOS+PFA-PBR-RGB~\cite{hu2022perspective} & 2022 & DNN & \cellcolor{ccol!100}Dataset & Extended FCOS & \cellcolor{ccol!100}PFA & \cellcolor{ccol!100}RGB & \cellcolor{ccol!100}PBR & \cellcolor{ccol!100}RGB & \cellcolor{arcol!74.5}74.5	&\cellcolor{arcol!71.9}71.9	&\cellcolor{arcol!73.2}73.2	&\cellcolor{arcol!60.0}60.0&	\cellcolor{arcol!35.3}35.3&	\cellcolor{arcol!84.1}84.1&	\cellcolor{arcol!64.8}64.8& \cellcolor{avgcol!66.3}66.3 & \cellcolor{timecol!34.97}$\phantom{0}$3.50 \\ 
        
        36 & PFA-cosypose~\cite{hu2022perspective,labbe2020cosypose}  & 2022 & DNN & \cellcolor{ccol!100}Dataset & MaskRCNN & \cellcolor{ccol!100}PFA & RGB-D & PBR+real & \cellcolor{ccol!100}RGB  & \cellcolor{arcol!67.4}67.4	&\cellcolor{arcol!73.8}73.8	&\cellcolor{arcol!83.7}83.7	&\cellcolor{arcol!59.6}59.6&	\cellcolor{arcol!24.6}24.6&	\cellcolor{arcol!71.2}71.2&	\cellcolor{arcol!80.7}80.7& \cellcolor{avgcol!65.9}65.9& $\phantom{0}$-- \\ 
         37 & Megapose-GDRNPPDet\_PBRReal~\cite{megapose} & 2022 & DNN & \cellcolor{ccol!100}Dataset & \cellcolor{ccol!100}Default GDRNPPDet & \cellcolor{ccol!100}~DeepIM & RGB & -- & \cellcolor{ccol!100}RGB &  \cellcolor{arcol!61.2}61.2 & \cellcolor{arcol!76.6}76.6 & \cellcolor{arcol!72.3}72.3 & \cellcolor{arcol!55.5}55.5 & \cellcolor{arcol!40.2}40.2 & \cellcolor{arcol!85.1}85.1 & \cellcolor{arcol!69.2}69.2 & 
        \cellcolor{avgcol!65.7}65.7 & \cellcolor{timecol!100}32.35 \\ 
        
        38 & SurfEmb-PBR-RGB~\cite{haugaard2022surfemb} & 2022 & DNN & \cellcolor{ccol!100}Dataset & \cellcolor{ccol!100}Default (PBR) & \cellcolor{ccol!100}Custom & \cellcolor{ccol!100}RGB & \cellcolor{ccol!100}PBR & \cellcolor{ccol!100}RGB & \cellcolor{arcol!66.3}66.3	&\cellcolor{arcol!73.5}73.5	&\cellcolor{arcol!71.5}71.5	&\cellcolor{arcol!58.8}58.8&	\cellcolor{arcol!41.3}41.3&	\cellcolor{arcol!79.1}79.1&	\cellcolor{arcol!64.7}64.7& \cellcolor{avgcol!65.0}65.0 & \cellcolor{timecol!88.91}$\phantom{0}$8.89 \\ 
        
        39 & Koenig-Hybrid-DL-PointPairs~\cite{koenig2020hybrid} & 2020 & DNN/PPF \cellcolor{ccol!100}& \cellcolor{ccol!100}Dataset & Retina/MaskRCNN & \cellcolor{ccol!100}{\raggedright ICP} & \cellcolor{ccol!100}RGB & Synt+real & RGB-D & \cellcolor{arcol!63.1}63.1	&\cellcolor{arcol!65.5}65.5	&\cellcolor{arcol!92.0}92.0	&\cellcolor{arcol!43.0}43.0&	\cellcolor{arcol!48.3}48.3&	\cellcolor{arcol!65.1}65.1&	\cellcolor{arcol!70.1}70.1& \cellcolor{avgcol!63.9}63.9 & \cellcolor{timecol!06.33}$\phantom{0}$0.63 \\
         40& CosyPose-ECCV20-SYNT+REAL-1VIEW~\cite{labbe2020cosypose} & 2020 & DNN & \cellcolor{ccol!100}Dataset & \cellcolor{ccol!100}Default (synt+real) & \cellcolor{ccol!100}${\tiny\sim}$DeepIM & \cellcolor{ccol!100}RGB & PBR+real & \cellcolor{ccol!100}RGB & \cellcolor{arcol!63.3}63.3	&\cellcolor{arcol!72.8}72.8	&\cellcolor{arcol!82.3}82.3	&\cellcolor{arcol!58.3}58.3&	\cellcolor{arcol!21.6}21.6&	\cellcolor{arcol!65.6}65.6&	\cellcolor{arcol!82.1}82.1& \cellcolor{avgcol!63.7}63.7 & \cellcolor{timecol!04.49}$\phantom{0}$0.45 \\ 
         
         41 & CRT-6D & 2022 & DNN & \cellcolor{ccol!100}Dataset & \cellcolor{ccol!100}Default (synt+real) & \cellcolor{ccol!100}Custom & \cellcolor{ccol!100}RGB & PBR+real & \cellcolor{ccol!100}RGB  & \cellcolor{arcol!66.0}66.0	&\cellcolor{arcol!64.4}64.4	&\cellcolor{arcol!78.9}78.9	&\cellcolor{arcol!53.7}53.7&	\cellcolor{arcol!20.8}20.8&	\cellcolor{arcol!60.3}60.3&	\cellcolor{arcol!75.2}75.2& \cellcolor{avgcol!59.9}59.9 & \cellcolor{timecol!0.059}$\phantom{0}$0.06 \\ 
         
         42 & Pix2Pose-BOP20\_w/ICP-ICCV19~\cite{park2019pix2pose} & 2020 & DNN & Object & MaskRCNN & \cellcolor{ccol!100}ICP & \cellcolor{ccol!100}RGB & PBR+real & RGB-D  & \cellcolor{arcol!58.8}58.8	&\cellcolor{arcol!51.2}51.2	&\cellcolor{arcol!82.0}82.0	&\cellcolor{arcol!39.0}39.0&	\cellcolor{arcol!35.1}35.1&	\cellcolor{arcol!69.5}69.5&	\cellcolor{arcol!78.0}78.0& \cellcolor{avgcol!59.1}59.1 & \cellcolor{timecol!48.44}$\phantom{0}$4.84 \\ 
         
         43 & ZTE\_PPF & 2022 & DNN/PPF \cellcolor{ccol!100}& \cellcolor{ccol!100}Dataset & \cellcolor{ccol!100}Default (synt+real) & \cellcolor{ccol!100}ICP & \cellcolor{ccol!100}RGB & PBR+real & RGB-D 
         & \cellcolor{arcol!66.3}66.3	&\cellcolor{arcol!37.4}37.4	&\cellcolor{arcol!90.4}90.4	&\cellcolor{arcol!39.6}39.6&	\cellcolor{arcol!47.0}47.0&	\cellcolor{arcol!73.5}73.5&	\cellcolor{arcol!50.2}50.2& \cellcolor{avgcol!57.8}57.8 & \cellcolor{timecol!0.901}$\phantom{0}$0.90 \\ 
 
         44 & CosyPose-ECCV20-PBR-1VIEW~\cite{labbe2020cosypose} & 2020 & DNN & \cellcolor{ccol!100}Dataset & \cellcolor{ccol!100}Default (PBR) & \cellcolor{ccol!100}${\tiny\sim}$DeepIM & \cellcolor{ccol!100}RGB & \cellcolor{ccol!100}PBR & \cellcolor{ccol!100}RGB & \cellcolor{arcol!63.3}63.3	&\cellcolor{arcol!64.0}64.0	&\cellcolor{arcol!68.5}68.5	&\cellcolor{arcol!58.3}58.3&	\cellcolor{arcol!21.6}21.6&	\cellcolor{arcol!65.6}65.6&	\cellcolor{arcol!57.4}57.4& \cellcolor{avgcol!57.0}57.0 & \cellcolor{timecol!04.75}$\phantom{0}$0.48 \\ 
         
         45 & Vidal-Sensors18~\cite{vidal2018method} & 2019 & PPF \cellcolor{ccol!100}& -- & -- & \cellcolor{ccol!100}ICP & -- & -- & D & \cellcolor{arcol!58.2}58.2	&\cellcolor{arcol!53.8}53.8	&\cellcolor{arcol!87.6}87.6	&\cellcolor{arcol!39.3}39.3&	\cellcolor{arcol!43.5}43.5&	\cellcolor{arcol!70.6}70.6&	\cellcolor{arcol!45.0}45.0& \cellcolor{avgcol!56.9}56.9 & \cellcolor{timecol!32.20}$\phantom{0}$3.22 \\ 
         
         46 & CDPNv2\_BOP20 (RGB-only \& ICP)~\cite{li2019cdpn} & 2020 & DNN & Object & FCOS & \cellcolor{ccol!100}ICP & \cellcolor{ccol!100}RGB & Synt+real & RGB-D & \cellcolor{arcol!63.0}63.0	&\cellcolor{arcol!46.4}46.4	&\cellcolor{arcol!91.3}91.3	&\cellcolor{arcol!45.0}45.0&	\cellcolor{arcol!18.6}18.6&	\cellcolor{arcol!71.2}71.2&	\cellcolor{arcol!61.9}61.9& \cellcolor{avgcol!56.8}56.8 & \cellcolor{timecol!14.62}$\phantom{0}$1.46 \\ 
         
         47 & Drost-CVPR10-Edges~\cite{drost2010model} & 2019 & PPF \cellcolor{ccol!100}& -- & -- & \cellcolor{ccol!100}ICP & -- & -- & RGB-D & \cellcolor{arcol!51.5}51.5	&\cellcolor{arcol!50.0}50.0	&\cellcolor{arcol!85.1}85.1	&\cellcolor{arcol!36.8}36.8&	\cellcolor{arcol!57.0}57.0&	\cellcolor{arcol!67.1}67.1&	\cellcolor{arcol!37.5}37.5& \cellcolor{avgcol!55.0}55.0 & \cellcolor{timecol!100}87.57 \\
         
         48 & MRPE-PBR-RGB-SModel & 2023 & \cellcolor{ccol!100}DNN & \cellcolor{ccol!100}Dataset & \cellcolor{ccol!100}Default GDRNPPDet & -- & RGB & PBR & RGB &  
        \cellcolor{arcol!71.5}71.5 & \cellcolor{arcol!72.9}72.9 & \cellcolor{arcol!20.6}20.6 & \cellcolor{arcol!46.2}46.2 & \cellcolor{arcol!35.3}35.3 & \cellcolor{arcol!76.5}76.5 & \cellcolor{arcol!55.2}55.2 & 
        \cellcolor{avgcol!54.0}54.0 & \cellcolor{timecol!10.0}$\phantom{0}$0.10 \\
        
        49 & CDPNv2\_BOP20 (PBR-only \& ICP)~\cite{li2019cdpn}  & 2020 & DNN & Object & FCOS & \cellcolor{ccol!100}ICP & \cellcolor{ccol!100}RGB & \cellcolor{ccol!100}PBR & RGB-D  & \cellcolor{arcol!63.0}63.0	&\cellcolor{arcol!43.5}43.5	&\cellcolor{arcol!79.1}79.1	&\cellcolor{arcol!45.0}45.0&	\cellcolor{arcol!18.6}18.6&	\cellcolor{arcol!71.2}71.2&	\cellcolor{arcol!53.2}53.2& \cellcolor{avgcol!53.4}53.4 & \cellcolor{timecol!14.91}$\phantom{0}$1.49 \\ 
         50 & CDPNv2\_BOP20 (RGB-only)~\cite{li2019cdpn} & 2020 & DNN & Object & FCOS & -- & \cellcolor{ccol!100}RGB & Synt+real & \cellcolor{ccol!100}RGB  & \cellcolor{arcol!62.4}62.4	&\cellcolor{arcol!47.8}47.8	&\cellcolor{arcol!77.2}77.2	&\cellcolor{arcol!47.3}47.3&	\cellcolor{arcol!10.2}10.2&	\cellcolor{arcol!72.2}72.2&	\cellcolor{arcol!53.2}53.2& \cellcolor{avgcol!52.9}52.9 & \cellcolor{timecol!09.35}$\phantom{0}$0.94 \\ 
         
         51 & Drost-CVPR10-3D-Edges~\cite{drost2010model} & 2019 & PPF \cellcolor{ccol!100}& -- & -- & \cellcolor{ccol!100}ICP & -- & -- & D & \cellcolor{arcol!46.9}46.9	&\cellcolor{arcol!40.4}40.4	&\cellcolor{arcol!85.2}85.2	&\cellcolor{arcol!37.3}37.3&	\cellcolor{arcol!46.2}46.2&	\cellcolor{arcol!62.3}62.3&	\cellcolor{arcol!31.6}31.6& \cellcolor{avgcol!50.0}50.0 & \cellcolor{timecol!100}80.06 \\ 
         
         52 & Drost-CVPR10-3D-Only~\cite{drost2010model} & 2019 & PPF \cellcolor{ccol!100}& -- & -- & \cellcolor{ccol!100}ICP & -- & -- & D & \cellcolor{arcol!52.7}52.7	&\cellcolor{arcol!44.4}44.4	&\cellcolor{arcol!77.5}77.5	&\cellcolor{arcol!38.8}38.8&	\cellcolor{arcol!31.6}31.6&	\cellcolor{arcol!61.5}61.5&	\cellcolor{arcol!34.4}34.4& \cellcolor{avgcol!48.7}48.7 & \cellcolor{timecol!77.04}$\phantom{0}$7.70 \\ 
         
         53 & CDPN\_BOP19 (RGB-only)~\cite{li2019cdpn} & 2020 & DNN & Object & RetinaNet & -- & \cellcolor{ccol!100}RGB & Synt+real & \cellcolor{ccol!100}RGB  & \cellcolor{arcol!56.9}56.9	&\cellcolor{arcol!49.0}49.0	&\cellcolor{arcol!76.9}76.9	&\cellcolor{arcol!32.7}32.7 & $\phantom{0}$\cellcolor{arcol!6.7}6.7&	\cellcolor{arcol!67.2}67.2&	\cellcolor{arcol!45.7}45.7& \cellcolor{avgcol!47.9}47.9 & \cellcolor{timecol!04.80}$\phantom{0}$0.48 \\ 
         
         54 & CDPNv2\_BOP20 (PBR-only \& RGB-only)~\cite{li2019cdpn}  & 2020 & DNN & Object & FCOS & -- & \cellcolor{ccol!100}RGB & \cellcolor{ccol!100}PBR & \cellcolor{ccol!100}RGB  & \cellcolor{arcol!62.4}62.4	&\cellcolor{arcol!40.7}40.7	&\cellcolor{arcol!58.8}58.8	&\cellcolor{arcol!47.3}47.3&	\cellcolor{arcol!10.2}10.2&	\cellcolor{arcol!72.2}72.2&	\cellcolor{arcol!39.0}39.0& \cellcolor{avgcol!47.2}47.2 & \cellcolor{timecol!09.78}$\phantom{0}$0.98 \\ 
         
         55 & leaping from 2D to 6D~\cite{liu2010leaping}& 2020 & DNN & Object & Unknown & -- & \cellcolor{ccol!100}RGB & Synt+real & \cellcolor{ccol!100}RGB   & \cellcolor{arcol!52.5}52.5	&\cellcolor{arcol!40.3}40.3	&\cellcolor{arcol!75.1}75.1	&\cellcolor{arcol!34.2}34.2&	$\phantom{0}$\cellcolor{arcol!7.7}7.7&	\cellcolor{arcol!65.8}65.8&	\cellcolor{arcol!54.3}54.3& \cellcolor{avgcol!47.1}47.1 & \cellcolor{timecol!04.25}$\phantom{0}$0.43 \\ 
         
         56 & EPOS-BOP20-PBR ~\cite{hodan2020epos} & 2020 & DNN & \cellcolor{ccol!100}Dataset & -- & -- & \cellcolor{ccol!100}RGB & \cellcolor{ccol!100}PBR & \cellcolor{ccol!100}RGB & \cellcolor{arcol!54.7}54.7	&\cellcolor{arcol!46.7}46.7	&\cellcolor{arcol!55.8}55.8	&\cellcolor{arcol!36.3}36.3&	\cellcolor{arcol!18.6}18.6&	\cellcolor{arcol!58.0}58.0&	\cellcolor{arcol!49.9}49.9& \cellcolor{avgcol!45.7}45.7 & \cellcolor{timecol!18.74}$\phantom{0}$1.87 \\ 
         
         57 & Drost-CVPR10-3D-Only-Faster~\cite{drost2010model} & 2019 & PPF \cellcolor{ccol!100}& -- & -- & \cellcolor{ccol!100}ICP & -- & -- & D & \cellcolor{arcol!49.2}49.2	&\cellcolor{arcol!40.5}40.5	&\cellcolor{arcol!69.6}69.6	&\cellcolor{arcol!37.7}37.7&	\cellcolor{arcol!27.4}27.4&	\cellcolor{arcol!60.3}60.3&	\cellcolor{arcol!33.0}33.0& \cellcolor{avgcol!45.4}45.4 & \cellcolor{timecol!13.83}$\phantom{0}$1.38 \\ 
         
         58 & Félix\&Neves-ICRA2017-IET2019~\cite{rodrigues2019deep,raposo2017using} & 2019 & DNN/PPF \cellcolor{ccol!100}& \cellcolor{ccol!100}Dataset & MaskRCNN & \cellcolor{ccol!100}ICP & RGB-D & Synt+real & RGB-D & \cellcolor{arcol!39.4}39.4	&\cellcolor{arcol!21.2}21.2	&\cellcolor{arcol!85.1}85.1	&\cellcolor{arcol!32.3}32.3&	$\phantom{0}$\cellcolor{arcol!6.9}6.9&	\cellcolor{arcol!52.9}52.9&	\cellcolor{arcol!51.0}51.0& \cellcolor{avgcol!41.2}41.2 & \cellcolor{timecol!100}55.78 \\ 
         
         59 & Sundermeyer-IJCV19+ICP~\cite{sundermeyer2019augmented} & 2019 & DNN & Object & RetinaNet & \cellcolor{ccol!100}ICP & \cellcolor{ccol!100}RGB & Synt+real & RGB-D & \cellcolor{arcol!23.7}23.7	&\cellcolor{arcol!48.7}48.7	&\cellcolor{arcol!61.4}61.4	&\cellcolor{arcol!28.1}28.1&	\cellcolor{arcol!15.8}15.8&	\cellcolor{arcol!50.6}50.6&	\cellcolor{arcol!50.5}50.5& \cellcolor{avgcol!39.8}39.8 & \cellcolor{timecol!08.65}$\phantom{0}$0.86 \\ 
         
         60& Zhigang-CDPN-ICCV19~\cite{li2019cdpn} & 2019 & DNN & Object & RetinaNet & -- & \cellcolor{ccol!100}RGB & Synt+real & \cellcolor{ccol!100}RGB & \cellcolor{arcol!37.4}37.4	&\cellcolor{arcol!12.4}12.4	&\cellcolor{arcol!75.7}75.7	&\cellcolor{arcol!25.7}25.7&	$\phantom{0}$\cellcolor{arcol!7.0}7.0&	\cellcolor{arcol!47.0}47.0&	\cellcolor{arcol!42.2}42.2& \cellcolor{avgcol!35.3}35.3 & \cellcolor{timecol!05.13}$\phantom{0}$0.51 \\ 
         
         61 & PointVoteNet2~\cite{hagelskjaer2019pointposenet} & 2020 & DNN & Object & -- & \cellcolor{ccol!100}ICP & RGB-D & \cellcolor{ccol!100}PBR & RGB-D & \cellcolor{arcol!65.3}65.3	& $\phantom{0}$\cellcolor{arcol!0.4}0.4	&\cellcolor{arcol!67.3}67.3	& \cellcolor{arcol!26.4}26.4&	$\phantom{0}$\cellcolor{arcol!0.1}0.1&	\cellcolor{arcol!55.6}55.6&	\cellcolor{arcol!30.8}30.8& \cellcolor{avgcol!35.1}35.1& $\phantom{0}$-- \\ 
         
         62 & Pix2Pose-BOP20-ICCV19 ~\cite{park2019pix2pose} & 2020 & DNN & Object & MaskRCNN & -- & \cellcolor{ccol!100}RGB & PBR+real & \cellcolor{ccol!100}RGB & \cellcolor{arcol!36.3}36.3	&\cellcolor{arcol!34.4}34.4	&\cellcolor{arcol!42.0}42.0	&\cellcolor{arcol!22.6}22.6&	\cellcolor{arcol!13.4}13.4&	\cellcolor{arcol!44.6}44.6&	\cellcolor{arcol!45.7}45.7& \cellcolor{avgcol!34.2}34.2 & \cellcolor{timecol!12.15}$\phantom{0}$1.22 \\ 
         
         63 & Sundermeyer-IJCV19\cite{sundermeyer2019augmented} & 2019 & DNN & Object & RetinaNet & -- & \cellcolor{ccol!100}RGB & Synt+real & \cellcolor{ccol!100}RGB & \cellcolor{arcol!14.6}14.6	&\cellcolor{arcol!30.4}30.4	&\cellcolor{arcol!40.1}40.1	&\cellcolor{arcol!21.7}21.7&	\cellcolor{arcol!10.1}10.1&	\cellcolor{arcol!34.6}34.6&	\cellcolor{arcol!44.6}44.6& \cellcolor{avgcol!28.0}28.0 & \cellcolor{timecol!01.96}$\phantom{0}$0.20 \\ 
         
         64 & SingleMultiPathEncoder-CVPR20 \cite{sundermeyer2020multi} & 2020 & DNN & \cellcolor{ccol!100}All & MaskRCNN & -- & \cellcolor{ccol!100}RGB & Synt+real & \cellcolor{ccol!100}RGB & \cellcolor{arcol!21.7}21.7	&\cellcolor{arcol!31.0}31.0	&\cellcolor{arcol!33.4}33.4	&\cellcolor{arcol!17.5}17.5&	$\phantom{0}$\cellcolor{arcol!6.7}6.7&	\cellcolor{arcol!29.3}29.3&	\cellcolor{arcol!28.9}28.9& \cellcolor{avgcol!24.1}24.1 & \cellcolor{timecol!01.86}$\phantom{0}$0.19 \\ 
         
         65 & DPOD (synthetic) ~\cite{zakharov2019dpod} & 2019 & DNN & \cellcolor{ccol!100}Dataset & -- & -- & \cellcolor{ccol!100}RGB & Synt & \cellcolor{ccol!100}RGB & \cellcolor{arcol!16.9}16.9	& $\phantom{0}$\cellcolor{arcol!8.1}8.1	&\cellcolor{arcol!24.2}24.2	&\cellcolor{arcol!13.0}13.0&	$\phantom{0}$\cellcolor{arcol!0.0}0.0&	\cellcolor{arcol!28.6}28.6&	\cellcolor{arcol!22.2}22.2& \cellcolor{avgcol!16.1}16.1 & \cellcolor{timecol!02.31}$\phantom{0}$0.23 \\
        \bottomrule
  
    \end{tabularx}
    \caption{\textbf{6D localization of seen objects (Task 1) on the seven core datasets.}
    The methods are ranked by the $\text{AR}_C$ score which is the average of the per-dataset $\text{AR}_D$ scores defined in Sec.~\ref{sec:task1}. The last column shows the average image processing time in seconds, \ie, the average time to localize all objects in an image (measured on different computers by the participants). Column \emph{Year} is the year of submission, \emph{Type} indicates whether the method relies on deep neural networks (DNN's) or point pair features (PPF's), \emph{DNN per...}~shows how many DNN models were trained, \emph{Det./seg.}~is the object detection or segmentation method, \emph{Refinement} is the pose refinement method, \emph{Train im.}~and \emph{Test im.}~show image channels used at training and test time respectively, and \emph{Train im.~type} is the domain of training images. All test images are real.
    }
    \label{tab:task1_results}
    
    \vspace{6ex}
\end{table*}

\setlength{\tabcolsep}{2pt}
\begin{table*}[t!]
    \tiny
    \centering
    \begin{tabularx}{\linewidth}{rlllllllll YYYYYYYYY}
        \toprule
        \# & Method & Year & Type & DNN per & Det./seg. & Refinement & Train im. & ...type & Test im. & LM-O & T-LESS & TUD-L & IC-BIN & ITODD & HB & YCB-V & \mbox{$\text{AR}_C$} & Time  \\ 
        \midrule
        1 & GenFlow-MultiHypo16~\cite{genflow} & \cellcolor{ccol!100}2023 & \cellcolor{ccol!100} DNN & \cellcolor{ccol!100} All & \cellcolor{ccol!100} CNOS-fastSAM & \cellcolor{ccol!100}Recurrent Flow & RGB-D & \cellcolor{ccol!100}PBR & RGB-D  & \cellcolor{arcol!63.5}63.5 & \cellcolor{arcol!52.1}52.1 & \cellcolor{arcol!86.2}86.2 & \cellcolor{arcol!53.4}53.4 & \cellcolor{arcol!55.4}55.4 & \cellcolor{arcol!77.9}77.9 & \cellcolor{arcol!83.3}83.3 & 
        \cellcolor{avgcol!67.4}67.4 &
        \cellcolor{timecol!34.58}$\phantom{0}$34.58 \\
        2 & GenFlow-MultiHypo~\cite{genflow} & \cellcolor{ccol!100}2023 & \cellcolor{ccol!100} DNN & \cellcolor{ccol!100} All & \cellcolor{ccol!100} CNOS-fastSAM & \cellcolor{ccol!100}Recurrent Flow & RGB-D & \cellcolor{ccol!100}PBR & RGB-D &  \cellcolor{arcol!62.2}62.2 & \cellcolor{arcol!50.9}50.9 & \cellcolor{arcol!84.9}84.9 & \cellcolor{arcol!52.4}52.4 & \cellcolor{arcol!54.4}54.4 & \cellcolor{arcol!77.0}77.0 & \cellcolor{arcol!81.8}81.8 & 
        \cellcolor{avgcol!66.2}66.2 & \cellcolor{timecol!21.46}$\phantom{0}$21.46 \\
        3 & Megapose-CNOS+Multih\_Teaserpp-10~\cite{megapose,nguyen2023cnos} & \cellcolor{ccol!100}2023 & \cellcolor{ccol!100} DNN & \cellcolor{ccol!100} All & \cellcolor{ccol!100} CNOS-fastSAM & \cellcolor{ccol!100}MegaPose+Teaser++ & \cellcolor{ccol!100}RGB & \cellcolor{ccol!100}PBR & RGB-D & \cellcolor{arcol!62.6}62.6 & \cellcolor{arcol!48.7}48.7 & \cellcolor{arcol!85.1}85.1 & \cellcolor{arcol!46.7}46.7 & \cellcolor{arcol!46.8}46.8 & \cellcolor{arcol!73.0}73.0 & \cellcolor{arcol!76.4}76.4 & 
        \cellcolor{avgcol!62.8}62.8 & \cellcolor{timecol!100}141.97 \\
        4 & Megapose-CNOS+Multih\_Teaserpp-10~\cite{megapose,nguyen2023cnos} & \cellcolor{ccol!100}2023 & \cellcolor{ccol!100} DNN & \cellcolor{ccol!100} All & \cellcolor{ccol!100} CNOS-fastSAM & \cellcolor{ccol!100}MegaPose+Teaser++ & \cellcolor{ccol!100}RGB & \cellcolor{ccol!100}PBR & RGB-D &  \cellcolor{arcol!62.0}62.0 & \cellcolor{arcol!48.5}48.5 & \cellcolor{arcol!84.6}84.6 & \cellcolor{arcol!46.2}46.2 & \cellcolor{arcol!46.0}46.0 & \cellcolor{arcol!72.5}72.5 & \cellcolor{arcol!76.4}76.4 & 
        \cellcolor{avgcol!62.3}62.3 & \cellcolor{timecol!100.}116.56 \\
        5 & SAM6D-CNOSmask~\cite{lin2023sam,nguyen2023cnos} & \cellcolor{ccol!100}2023 & \cellcolor{ccol!100} DNN & \cellcolor{ccol!100} All & \cellcolor{ccol!100} CNOS-fastSAM & \cellcolor{ccol!100}Cross-attention & RGB-D & \cellcolor{ccol!100}PBR & RGB-D &   \cellcolor{arcol!64.8}64.8 & \cellcolor{arcol!48.3}48.3 & \cellcolor{arcol!79.4}79.4 & \cellcolor{arcol!50.4}50.4 & \cellcolor{arcol!35.1}35.1 & \cellcolor{arcol!72.7}72.7 & \cellcolor{arcol!80.4}80.4 & 
        \cellcolor{avgcol!61.6}61.6 & \cellcolor{timecol!3.872}$\phantom{0}\phantom{0}$3.87 \\
        6 & PoZe (CNOS) & \cellcolor{ccol!100}2023 & \cellcolor{ccol!100} DNN & \cellcolor{ccol!100} All & \cellcolor{ccol!100} CNOS-fastSAM & \cellcolor{ccol!100}ICP & RGB-D & Custom & RGB-D &  \cellcolor{arcol!64.4}64.4 & \cellcolor{arcol!49.4}49.4 & \cellcolor{arcol!92.4}92.4 & \cellcolor{arcol!40.9}40.9 & \cellcolor{arcol!51.6}51.6 & \cellcolor{arcol!71.2}71.2 & \cellcolor{arcol!61.1}61.1 & 
        \cellcolor{avgcol!61.6}61.6 &
        \cellcolor{timecol!100}159.43 \\
        7 & ZeroPose-Multi-Hypo-Refinement~\cite{chen20233d,nguyen2023cnos} & \cellcolor{ccol!100}2023 & \cellcolor{ccol!100} DNN & \cellcolor{ccol!100} All & \cellcolor{ccol!100} CNOS-fastSAM & \cellcolor{ccol!100}MegaPose & RGB-D & PBR+real & RGB-D &   \cellcolor{arcol!53.8}53.8 & \cellcolor{arcol!40.0}40.0 & \cellcolor{arcol!83.5}83.5 & \cellcolor{arcol!39.2}39.2 & \cellcolor{arcol!52.1}52.1 & \cellcolor{arcol!65.3}65.3 & \cellcolor{arcol!65.3}65.3 & 
        \cellcolor{avgcol!57.0}57.0 &
        \cellcolor{timecol!16.168}$\phantom{0}$16.17 \\
        8 & GenFlow-MultiHypo-RGB & \cellcolor{ccol!100}2023 & \cellcolor{ccol!100} DNN & \cellcolor{ccol!100} All & \cellcolor{ccol!100} CNOS-fastSAM & \cellcolor{ccol!100}Recurrent Flow & RGB-D & \cellcolor{ccol!100}PBR & \cellcolor{ccol!100}RGB &   \cellcolor{arcol!56.3}56.3 & \cellcolor{arcol!52.3}52.3 & \cellcolor{arcol!68.4}68.4 & \cellcolor{arcol!45.3}45.3 & \cellcolor{arcol!39.5}39.5 & \cellcolor{arcol!73.9}73.9 & \cellcolor{arcol!63.3}63.3 & 
        \cellcolor{avgcol!57.0}57.0 & \cellcolor{timecol!20.890}$\phantom{0}$20.89 \\
        9 & Megapose-CNOS\_fastSAM+Multih-10~\cite{megapose,nguyen2023cnos} & \cellcolor{ccol!100}2023 & \cellcolor{ccol!100} DNN & \cellcolor{ccol!100} All & \cellcolor{ccol!100} CNOS-fastSAM & \cellcolor{ccol!100}MegaPose & \cellcolor{ccol!100}RGB & \cellcolor{ccol!100}PBR & \cellcolor{ccol!100}RGB &    \cellcolor{arcol!56.0}56.0 & \cellcolor{arcol!50.8}50.8 & \cellcolor{arcol!68.7}68.7 & \cellcolor{arcol!41.9}41.9 & \cellcolor{arcol!34.6}34.6 & \cellcolor{arcol!70.6}70.6 & \cellcolor{arcol!62.0}62.0 & 
        \cellcolor{avgcol!54.9}54.9 &
        \cellcolor{timecol!53.878}$\phantom{0}$53.88 \\
        10 & Megapose-CNOS\_fastSAM+Multih~\cite{megapose,nguyen2023cnos}& \cellcolor{ccol!100}2023 & \cellcolor{ccol!100} DNN & \cellcolor{ccol!100} All & \cellcolor{ccol!100} CNOS-fastSAM & \cellcolor{ccol!100}MegaPose & \cellcolor{ccol!100}RGB & \cellcolor{ccol!100}PBR & \cellcolor{ccol!100}RGB  & \cellcolor{arcol!56.0}56.0 & \cellcolor{arcol!50.7}50.7 & \cellcolor{arcol!68.4}68.4 & \cellcolor{arcol!41.4}41.4 & \cellcolor{arcol!33.8}33.8 & \cellcolor{arcol!70.4}70.4 & \cellcolor{arcol!62.1}62.1 & 
        \cellcolor{avgcol!54.7}54.7 & 
        \cellcolor{timecol!47.386}$\phantom{0}$47.39 \\
        11 & ZeroPose-Multi-Hypo-Refinement~\cite{chen20233d,nguyen2023cnos}& \cellcolor{ccol!100}2023 & \cellcolor{ccol!100} DNN & \cellcolor{ccol!100} All & SAM + ImageBind & \cellcolor{ccol!100}MegaPose & RGB-D & PBR+real & RGB-D & \cellcolor{arcol!49.3}49.3 & \cellcolor{arcol!34.2}34.2 & \cellcolor{arcol!79.0}79.0 & \cellcolor{arcol!39.6}39.6 & \cellcolor{arcol!46.5}46.5 & \cellcolor{arcol!62.9}62.9 & \cellcolor{arcol!62.3}62.3 & 
        \cellcolor{avgcol!53.4}53.4 & 
        \cellcolor{timecol!18.971}$\phantom{0}$18.97 \\
        12 & MegaPose-CNOS\_fastSAM~\cite{megapose,nguyen2023cnos} & \cellcolor{ccol!100}2023 & \cellcolor{ccol!100} DNN & \cellcolor{ccol!100} All & \cellcolor{ccol!100} CNOS-fastSAM & \cellcolor{ccol!100}MegaPose & \cellcolor{ccol!100}RGB & \cellcolor{ccol!100}PBR & \cellcolor{ccol!100}RGB & \cellcolor{arcol!49.9}49.9 & \cellcolor{arcol!47.7}47.7 & \cellcolor{arcol!65.3}65.3 & \cellcolor{arcol!36.7}36.7 & \cellcolor{arcol!31.5}31.5 & \cellcolor{arcol!65.4}65.4 & \cellcolor{arcol!60.1}60.1 & 
        \cellcolor{avgcol!50.9}50.9 & \cellcolor{timecol!31.724}$\phantom{0}$31.72 \\
        13 & ZeroPose-One-Hypo~\cite{chen20233d}& \cellcolor{ccol!100}2023 & \cellcolor{ccol!100} DNN & \cellcolor{ccol!100} All & SAM + ImageBind & \cellcolor{ccol!100}MegaPose & RGB-D & PBR+real & RGB-D & \cellcolor{arcol!27.2}27.2 & \cellcolor{arcol!15.6}15.6 & \cellcolor{arcol!53.6}53.6 & \cellcolor{arcol!30.7}30.7 & \cellcolor{arcol!36.2}36.2 & \cellcolor{arcol!46.2}46.2 & \cellcolor{arcol!34.1}34.1 & 
        \cellcolor{avgcol!34.8}34.8 & \cellcolor{timecol!9.756}$\phantom{0}\phantom{0}$9.76 \\
        14 & GenFlow-coarse & \cellcolor{ccol!100}2023 & \cellcolor{ccol!100} DNN & \cellcolor{ccol!100} All & \cellcolor{ccol!100} CNOS-fastSAM  & -- & \cellcolor{ccol!100}RGB & \cellcolor{ccol!100}PBR & \cellcolor{ccol!100}RGB &  \cellcolor{arcol!25.0}25.0 & \cellcolor{arcol!21.5}21.5 & \cellcolor{arcol!30.0}30.0 & \cellcolor{arcol!16.8}16.8 & \cellcolor{arcol!15.4}15.4 & \cellcolor{arcol!28.3}28.3 & \cellcolor{arcol!27.7}27.7 & 
        \cellcolor{avgcol!23.5}23.5 & \cellcolor{timecol!3.839}$\phantom{0}\phantom{0}$3.84  \\
        \bottomrule
  
    \end{tabularx}
    \caption{\textbf{6D localization of unseen objects (Task 4) on the seven core datasets.}
    The methods are ranked by the $\text{AR}_C$ score which is the average of the per-dataset $\text{AR}_D$ scores defined in Sec.~\ref{sec:task4}. The last column shows the average image processing time (in seconds). Other columns as in Tab.~\ref{tab:task1_results}.
    }
    \label{tab:task4_results}
    
    \vspace{0.5ex}
\end{table*}

\setlength{\tabcolsep}{2pt}
\begin{table}[t]
    \tiny
    \begin{tabularx}{\columnwidth}{lLLlllll} \toprule
        \# & Method & ...based on & Year & Data & ...type & AP$_C$ & Time \\ 
        \midrule
        1 & GDet2023 & YOLOv8 & \cellcolor{ccol!100}2023 & \cellcolor{ccol!100}RGB & PBR+real &  \cellcolor{avgcol!79.8}79.8 & \cellcolor{timecol!41}.204 \\ 
        2 & GDRNPPDet & YOLOX & 2022 & \cellcolor{ccol!100}RGB & PBR+real &  \cellcolor{avgcol!77.3}77.3 & \cellcolor{timecol!41}.081 \\ 
        3 & GDet2023-PBR & YOLOv8 & \cellcolor{ccol!100}2023 & \cellcolor{ccol!100}RGB & \cellcolor{ccol!100}PBR &  \cellcolor{avgcol!76.9}76.9 & \cellcolor{timecol!41}.204 \\ 
        4 & GDRNPPDet & YOLOX & 2022 & \cellcolor{ccol!100}RGB & \cellcolor{ccol!100}PBR & \cellcolor{avgcol!73.8}73.8 & \cellcolor{timecol!41}.081 \\ 
        5 & Extended\_FCOS & FCOS & 2022 & \cellcolor{ccol!100}RGB & PBR+real & \cellcolor{avgcol!72.1}72.1 & \cellcolor{timecol!15}.030 \\ 
        6 & Extended\_FCOS & FCOS & 2022 & \cellcolor{ccol!100}RGB & \cellcolor{ccol!100}PBR & \cellcolor{avgcol!66.7}66.7 & \cellcolor{timecol!15}.030 \\ 
        7 & DLZDet & DLZDet & 2022 & \cellcolor{ccol!100}RGB & \cellcolor{ccol!100}PBR & \cellcolor{avgcol!65.6}65.6 & - \\ 
        8 & CosyPose & Mask R-CNN & 2020 & \cellcolor{ccol!100}RGB & PBR+real & \cellcolor{avgcol!60.5}60.5 & \cellcolor{timecol!27}.054 \\
        9 & CosyPose & Mask R-CNN & 2020 & \cellcolor{ccol!100}RGB & \cellcolor{ccol!100}PBR & \cellcolor{avgcol!55.7}55.7 & \cellcolor{timecol!28}.055 \\
        10 & FCOS-CDPN & FCOS & 2022 & \cellcolor{ccol!100}RGB & \cellcolor{ccol!100}PBR & \cellcolor{avgcol!50.7}50.7 & \cellcolor{timecol!24}.047 \\ 
        \bottomrule
    \end{tabularx}
    \caption{\textbf{2D detection of seen objects (Task 2).}
    The methods are ranked by the $\text{AP}_C$ score defined in Sec.~\ref{sec:task2}. The last column shows the average image processing time (in seconds).
    }
    \label{tab:det_results}
    \vspace{1.8ex}
\end{table}

\setlength{\tabcolsep}{2pt}
\begin{table}[t]
    \tiny
    \begin{tabularx}{\columnwidth}{lLLlllll} \toprule
        \# & Method & ...based on & Year & Data & ...type & AP$_C$ & Time \\ 
        \midrule
        1 & ZebraPoseSAT & GDRNPP+Zebra & \cellcolor{ccol!100}2023 & \cellcolor{ccol!100}RGB & PBR+real & \cellcolor{avgcol!61.9}61.9 & \cellcolor{timecol!40}.080 \\ 
        2 & ZebraPoseSAT & GDRNPP+Zebra & 2022 & \cellcolor{ccol!100}RGB & PBR+real & \cellcolor{avgcol!58.7}58.7 & \cellcolor{timecol!40}.080 \\ 
        3 & ZebraPoseSAT & CDPNv2+Zebra & \cellcolor{ccol!100}2023 & \cellcolor{ccol!100}RGB & PBR+real & \cellcolor{avgcol!57.9}57.9 & \cellcolor{timecol!40}.080 \\ 
        4 & ZebraPoseSAT & CDPNv2+Zebra & 2022 & \cellcolor{ccol!100}RGB & PBR+real & \cellcolor{avgcol!57.8}57.8 & \cellcolor{timecol!40}.080 \\ 
        5 & ZebraPoseSAT & CosyPose+Zebra & 2022 & \cellcolor{ccol!100}RGB & \cellcolor{ccol!100}PBR & \cellcolor{avgcol!53.8}53.8 & \cellcolor{timecol!40}.080 \\ 
        6 & ZebraPoseSAT & CDPNv2+Zebra & 2022 & \cellcolor{ccol!100}RGB & \cellcolor{ccol!100}PBR & \cellcolor{avgcol!52.3}52.3 & \cellcolor{timecol!40}.080 \\ 
        7 & DLZDet & DLZDet & 2022 & \cellcolor{ccol!100}RGB & PBR+real & \cellcolor{avgcol!49.6}49.6 & - \\ 
        8 & DLZDet & DLZDet & 2022 & \cellcolor{ccol!100}RGB & \cellcolor{ccol!100}PBR & \cellcolor{avgcol!42.9}42.9 & - \\ 
        9 & CosyPose & Mask R-CNN & 2020 & \cellcolor{ccol!100}RGB & PBR+real & \cellcolor{avgcol!40.5}40.5 & \cellcolor{timecol!27}.054 \\ 
        10 & CosyPose & Mask R-CNN & 2020 & \cellcolor{ccol!100}RGB & \cellcolor{ccol!100}PBR & \cellcolor{avgcol!36.2}36.2 & \cellcolor{timecol!28}.055 \\ 
        \bottomrule
    \end{tabularx}
    \caption{\textbf{2D segmentation of seen objects (Task 3).}
    Details as in Tab.~\ref{tab:det_results}.
    }
    \label{tab:seg_results}
    \vspace{0.5ex}
\end{table}

\setlength{\tabcolsep}{2pt}
\begin{table}[t!]
    \tiny
    \begin{tabularx}{\columnwidth}{lLllllll} \toprule
        \# & Method & Year & Train. im. & ...type & Test. im. & AP$_C$ & Time \\ 
        \midrule
        1 & CNOS\_FastSAM~\cite{nguyen2023cnos} & \cellcolor{ccol!100}2023 & \cellcolor{ccol!100}RGB & \cellcolor{ccol!100}PBR & 
        \cellcolor{ccol!100}RGB & \cellcolor{avgcol!42.8}42.8 & \cellcolor{timecol!2.21}0.221 \\ 
        2 & CNOS\_SAM~\cite{nguyen2023cnos} & \cellcolor{ccol!100}2023 & \cellcolor{ccol!100}RGB & \cellcolor{ccol!100}PBR & 
        \cellcolor{ccol!100}RGB & \cellcolor{avgcol!36.1}36.1 & \cellcolor{timecol!18.47}1.847 \\ 
        3 & ZeroPose~\cite{chen20233d} & \cellcolor{ccol!100}2023 & \cellcolor{ccol!100}RGB & \cellcolor{ccol!100}PBR &  
        \cellcolor{ccol!100}RGB & \cellcolor{avgcol!34.1}34.1 & \cellcolor{timecol!38.21}3.821 \\ 
        
        \bottomrule
    \end{tabularx}
    \caption{\textbf{2D detection of unseen objects (Task 5).}
    The methods are ranked by the $\text{AP}_C$ score defined in Sec.~\ref{sec:task5}. The last column shows the average image processing time (in seconds).
    }
    \label{tab:det_results_unseen}
    \vspace{1.8ex}
\end{table}

\setlength{\tabcolsep}{2pt}
\begin{table}[t!]
    \tiny
    \begin{tabularx}{\columnwidth}{lLllllll} \toprule
        \# & Method & Year & Train. im. & ...type & Test. im. & AP$_C$ & Time \\ 
        \midrule
        1 & CNOS\_FastSAM~\cite{nguyen2023cnos} & \cellcolor{ccol!100}2023 & \cellcolor{ccol!100}RGB & \cellcolor{ccol!100}PBR & 
        \cellcolor{ccol!100}RGB & \cellcolor{avgcol!41.2}41.2 & \cellcolor{timecol!2.21}0.221 \\ 
        2 & CNOS\_SAM~\cite{nguyen2023cnos} & \cellcolor{ccol!100}2023 & \cellcolor{ccol!100}RGB & \cellcolor{ccol!100}PBR & 
        \cellcolor{ccol!100}RGB & \cellcolor{avgcol!40.3}40.3 & \cellcolor{timecol!18.47}1.847 \\ 
        3 & ZeroPose~\cite{chen20233d} & \cellcolor{ccol!100}2023 & \cellcolor{ccol!100}RGB & \cellcolor{ccol!100}PBR &  
        \cellcolor{ccol!100}RGB & \cellcolor{avgcol!37.2}37.2 & \cellcolor{timecol!38.21}3.821 \\
        4 & lcc-fastsam & \cellcolor{ccol!100}2023 & \cellcolor{ccol!100}RGB & \cellcolor{ccol!100}PBR &  
        \cellcolor{ccol!100}RGB & \cellcolor{avgcol!14.9}14.9 & \cellcolor{timecol!11.82}1.182 \\
        \bottomrule
    \end{tabularx}
    \caption{\textbf{2D segment.\ of unseen objects (Task 6).}
    Details as in Tab.~\ref{tab:det_results_unseen}.
    }
    \label{tab:seg_results_unseen}
    \vspace{0.5ex}
\end{table}

\subsection{Results on Task 1} \label{sec:results_task_1}

Results on the task of 6D object localization of seen objects and properties of the evaluated methods are in Tab.~\ref{tab:task1_results}. Among the 16 new entries in 2023, three outperform GDRNPP~\cite{Wang_2021_GDRN, liu2022gdrnpp_bop}, the best method from the 2022 challenge.
The best pose estimation pipeline from 2023, GPose2023~\cite{Wang_2021_GDRN,gpose2023}, is purely learning-based and achieves 85.6 AR$_C$, outperforming GDRNPP by 1.9~AR$_C$ (\#1$-$\#4 in Tab.~\ref{tab:task1_results}) with less than half the inference time (2.67\,s vs. 6.26\,s). 
GPose2023 deploys the same pose estimation method as GDRNPP but combines it with a more efficient coordinate-guided pose refinement strategy~\cite{gpose2023} and an improved 2D object detector based on YOLOv8 (see \#1$-$\#2 in Tab.~\ref{tab:det_results}). 
Without any pose refinement, the RGB-only variants GPose2023-RGB (\#21, 72.9~AR$_C$) or ZebraPoseSAT-EffnetB4~\cite{su2022zebrapose} (\#17, 74.9~AR$_C$) reach an average inference time of $\sim$0.25 seconds per image which are closer to the demands of mobile vision applications.
Gains in accuracy are most notable on the industrial ITODD, T-LESS, and HB datasets, whereas on TUD-L and YCB-V we can observe that metrics start to saturate.

\subsection{Results on Tasks 2 and 3} \label{sec:results_tasks_2_3}

As shown in Tab.~\ref{tab:det_results}, GDet2023~\cite{gpose2023} based on YOLOv8~\cite{Jocher_Ultralytics_YOLO_2023} achieves 79.8 AP$_C$, a moderate +2.5~AP$_C$ gain over YOLOX~\cite{ge2021yolox}, the best detector in 2022. YOLOv8 is even less sensitive to the training image domain than YOLOX, achieving 76.9~AP$_C$ when trained only on synthetic PBR images and neglecting the real training data.
In the 2D segmentation of seen objects task (Tab.~\ref{tab:seg_results}), we see a similar incremental improvement of +3.2~AP$_C$ achieved by ZebraPoseSAT~\cite{su2022zebrapose}, which predicts object masks from the provided default detections of GDRNPP$\_$Det.

\subsection{Results on Task 4} \label{sec:results_task_4}

The new task of 6D localization of unseen objects received 14 entries, as presented in Tab.~\ref{tab:task4_results}. 
MegaPose~\cite{megapose}, a method from 2022, was considered as the baseline and consists of two stages: (1) coarse object pose estimation by finding the rendered template image that is closest to the test image crop, and (2) pose refinement via a render-and-compare strategy.
The RGB-only entry Megapose-CNOS\_fastSAM+Multih-10 (\#9) achieves 54.9~AR$_C$ and further improves to 62.8 AR$_C$ by using RGB-D images and an additional refinement with Teaser++~\cite{teaserpp}, see Megapose-CNOS+Multih\_Teaserpp-10 (\#3). 

GenFlow-MultiHypo16 (\#1), the best method for 6D localization of unseen objects, reaches 67.4 AR$_C$. 
This is a remarkable result since the performance is comparable to CosyPose~\cite{labbe2020cosypose}, the best method in 6D localization of seen objects from 2020. 
GenFlow improves the coarse pose estimation stage of MegaPose by running the coarse network in a GMM-based hierarchical manner.
For pose refinement, GenFlow adapts the recurrent flow network~\cite{hai2023shape} to also estimate a visibility mask and replaces the pose regression network with a differentiable P\emph{n}P solver.

Results in Tab.~\ref{tab:task4_results} highlight that the run time is a significant challenge for solving unseen object pose localization.
While GenFlow-MultiHypo16 improved the run time by 4x compared to MegaPose, it still takes 34.58\,s per image. SAM6D (\#5)~\cite{lin2023sam} based on GeoTransformer~\cite{qin2023geotransformer} is the fastest method by a significant margin with 3.87\,s per image while still reaching 61.6 AR$_C$ (-5.8 AR$_C$ compared to Genflow-MultiHypo16 \#1).
Figure~\ref{fig:cover_results} shows qualitative comparison of the best method for unseen objects, GenFlow, with the best method for seen objects, GPose.

\subsection{Results on Tasks 5 and 6} \label{sec:results_tasks_5_6}

2D detection and segmentation of unseen objects in cluttered, occluded environments is a challenging task.
Still, as shown in Tab.~\ref{tab:det_results_unseen} and Tab.~\ref{tab:seg_results_unseen}, the best method CNOS-FastSAM~\cite{nguyen2023cnos} reaches accuracy of 42.8~mAP$_C$ in detection and 41.2~mAP in segmentation of unseen objects.
For comparison, the instance segmentation accuracy is comparable to Mask R-CNN~\cite{he2017mask} that reached 40.5~mAP$_C$ in the BOP challenge 2020~\cite{hodan2020bop} while being trained on more than 1M synthetic and real images of the target objects.
CNOS-FastSAM~\cite{nguyen2023cnos} instead relies on DINOv2~\cite{oquab2023dinov2} features extracted from only 200 rendered reference views per object.
All submitted detection and segmentation approaches are RGB-based and rely on SAM-like (Segment Anything)~\cite{lin2023sam} methods to segment object instances in the image.

Despite the substantial progress in unseen object detection and segmentation driven by foundation models, there is still a relatively large gap to methods trained to detect and segment specific objects (compare Tab.~\ref{tab:det_results} and \ref{tab:seg_results}).
Especially, the amodal detection of occluded instances, \ie, including occluded parts, is a clear challenge for approaches focusing on unseen objects, leading to a gap of 37~mAP$_C$ between CNOS and GDet2023.

To what extent is this gap in 2D detection performance responsible for the gap in 6D localization of seen and unseen objects?
When combined with the default GDRNPPDet detections from Task 2, the best method for 6D localization of unseen objects (GenFlow-MultiHypo16) achieves the pose accuracy of 79.2 AR$_C$ (\#10 Tab.~\ref{tab:task1_results}). Since this is only 5.9 AR$_C$ behind GDRNPPDet + GPose2023 (\#2), we conclude that better methods for unseen object detection would provide great potential for improving methods for unseen object localization.

\section{Awards} \label{sec:awards}
The BOP Challenge 2023 awards were presented at the 8th Workshop on Recovering 6D Object Pose\footnote{
\texttt{\href{https://cmp.felk.cvut.cz/sixd/workshop_2023/}{cmp.felk.cvut.cz/sixd/workshop\_2023}}} at the ICCV 2023 conference. The awards are based on the results analyzed in Sec.~\ref{sec:evaluation}. The submissions were prepared by the following authors:
\begin{itemize}
    \item GPose2023 and GDet2023~\cite{gpose2023} by Ruida Zhang, Ziqin Huang, Gu Wang, Xingyu Liu, Chenyangguang Zhang, Xiangyang Ji
    \item GDRNPP~\cite{Wang_2021_GDRN, liu2022gdrnpp_bop} by Xingyu Liu, Ruida Zhang, Chenyangguang Zhang, Bowen Fu, Jiwen Tang, Xiquan Liang, Jingyi Tang, Xiaotian Cheng, Yukang Zhang, Gu Wang, Xiangyang Ji
    \item OfficialDet-PFA~\cite{hu2022perspective} by Xinyao Fan, Fengda Hao, Yang Hai, Jiaojiao Li, Rui Song, Haixin Shi, Mathieu Salzmann, David Ferstl, Yinlin Hu
    \item ZebraPoseSAT~\cite{su2022zebrapose} by Praveen Annamalai Nathan, Sandeep Prudhvi Krishna Inuganti, Yongliang Lin, Yongzhi Su,Yu Zhang, Didier Stricker, Jason Rambach
    \item Coupled Iterative Refinement~\cite{lipson2022coupled} by Lahav Lipson, Zachary Teed, Ankit Goyal, Jia Deng
    \item GenFlow~\cite{genflow} by Sungphill Moon, Hyeontae Son.
    \item SAM6D~\cite{lin2023sam} by Jiehong Lin, Lihua Liu, Dekun Lu, Kui Jia
    \item MegaPose~\cite{megapose} by Elliot Maitre, Mederic Fourmy, Lucas Manuelli, Yann Labb{\'e}
    \item PoZe by Andrea Caraffa, Davide Boscaini, Fabio Poiesi
    \item CNOS~\cite{nguyen2023cnos} by Van Nguyen Nguyen, Thibault Groueix, Georgy Ponimatkin, Vincent Lepetit, Tomas Hodan
\end{itemize}
\vspace{1.0em}
\noindent Awards for 6D localization of seen objects (Task 1):
\vspace{0.5em}
\begin{itemize}
    \item \textbf{The Overall Best Method:} \\ GPose2023
    \item \textbf{The Best RGB-Only Method:} \\ ZebraPoseSAT-EffnetB4
    \item \textbf{The Best Fast Method (less than 1s per image):} \\ GDRNPP-PBRReal-RGBD-MModel-Fast
    \item \textbf{The Best BlenderProc-Trained Method:} \\ GPose2023-PBR
    \item \textbf{The Best Single-Model Method} (trained per dataset)\textbf{:} \\ OfficialDet-PFA-Mixpbr-RGB-D
    \item \textbf{The Best Open-Source Method:} \\ GDRNPP-PBRReal-RGBD-MModel
    \item \textbf{The Best Method Using Default Detections:} \\ GPose2023-OfficialDet
    \item \textbf{The Best Method on T-LESS, ITODD, HB, IC-BIN:} \\ GPose2023
    \item \textbf{The Best Method on LM-O, YCB-V:} \\ GPose2023-OfficialDet
    \item \textbf{The Best Method on TUD-L:} \\ Coupled Iterative Refinement (CIR)
\end{itemize}

\vspace{1.0em}
\noindent Awards for 2D detect./segment.\ of seen objects (Tasks 2 and 3):
\vspace{0.5em}
\begin{itemize}
    \item \textbf{The Overall Best Detection Method:} \\ GDet2023
    \item \textbf{The Best BlenderProc-Trained Detection Method:} \\ GDet2023-PBR   

    \item \textbf{The Overall Best Segmentation Method:} \\ ZebraPoseSAT-EffnetB4 (DefaultDetection)
    \item \textbf{The Best BlenderProc-Trained Segment. Method:} \\ ZebraPoseSAT-EffnetB4 (DefaultDet+PBR\_Only)

    \setlength\itemsep{-0.3em}
\end{itemize}

\vspace{1.0em}
\noindent Awards for 6D localization of unseen objects (Task 4):
\vspace{0.5em}
\begin{itemize}
    \item \textbf{The Overall Best Method:} \\ GenFlow-MultiHypo16
    \item \textbf{The Best RGB-Only Method:} \\ GenFlow-MultiHypo-RGB
    \item \textbf{The Best Fast Method (less than 1s per image):} \\ SAM6D-CNOSmask
    \item \textbf{The Best BlenderProc-Trained Method:} \\ GenFlow-MultiHypo16
    \item \textbf{The Best Single-Model Method (one for all core datasets)} \textbf{:} \\ GenFlow-MultiHypo16
    \item \textbf{The Best Open-Source Method:} \\ Megapose-CNOS\_fastSAM+Multih\_Teaserpp-10
    \item \textbf{The Best Method Using Default Detections/Segmentations:} \\ GenFlow-MultiHypo16
    \item \textbf{The Best Method on ITODD, IC-BIN, HB, YCB-V:} \\ GenFlow-MultiHypo16
    \item \textbf{The Best Method on T-LESS:} \\ GenFlow-MultiHypo-RGB
    \item \textbf{The Best Method on LM-O:} \\ SAM6D-CNOSmask
    \item \textbf{The Best Method on TUD-L:} \\ PoZe (CNOS)
\end{itemize}

\vspace{1.0em}
\noindent Awards for 2D detect./segment.\ of unseen objects (Tasks 5 and 6):
\vspace{0.5em}
\begin{itemize}
    \item \textbf{The Overall Best Detection Method:} \\ CNOS (FastSAM)
    \item \textbf{The Best BlenderProc-Trained Detection Method:} \\ CNOS (FastSAM)  

    \item \textbf{The Overall Best Segmentation Method:} \\ CNOS (FastSAM)
    \item \textbf{The Best BlenderProc-Trained Segment. Method:} \\ CNOS (FastSAM)

    \setlength\itemsep{-0.3em}
\end{itemize}

\section{Conclusions} \label{sec:conclusion}
Although the accuracy scores start saturating on the seen-object tasks (Tasks 1--3), the top-performing methods still need to improve efficiency in order to support real-time applications. 2023 was a strong first year for the new unseen-object tasks (Tasks 4--6), with the top performing method for 6D localization of unseen objects reaching the accuracy of the top 2020 method for 6D localization of seen objects. However, we identified a great potential in improving detection of occluded objects and making unseen object pose estimation more efficient. In 2023, methods for unseen objects were provided 3D mesh models to onboard the target objects. Next years, we are planning to introduce an even more challenging variant where only reference images of each object are provided for the onboarding.
The evaluation system at\ \texttt{\href{http://bop.felk.cvut.cz/}{bop.felk.cvut.cz}}\ stays open and raw results of all methods are publicly available.

{\small
\bibliographystyle{ieee_fullname}
\bibliography{references}
}

\end{document}